\documentclass[runningheads]{llncs}
\usepackage{graphicx}
\usepackage{amsmath,amssymb} % define this before the line numbering.
\usepackage{color}
\usepackage[width=122mm,left=12mm,paperwidth=146mm,height=193mm,top=12mm,paperheight=217mm]{geometry}

\usepackage{times}
\usepackage{epsfig}
\usepackage{subfigure}
\usepackage{esvect}
\usepackage{amsmath}
\usepackage{mathtools}
\usepackage{mathrsfs}
\usepackage{kotex}
\usepackage[switch]{lineno}
\usepackage{booktabs}
\usepackage{url}
\usepackage{hyperref}
\usepackage{cleveref}
\usepackage[export]{adjustbox}
\usepackage{multirow}
\usepackage{tabu}
\usepackage{boldline}
\usepackage{algorithm}
\usepackage{algpseudocode}
\usepackage{etoolbox}
\usepackage{float}
\usepackage{threeparttable}

\newcommand\tabgab{-0.4ex} % by ymlee.
\newcommand\cellgab{1.0ex} % by ymlee.
\newcommand\cellgabb{0.8ex} % by ymlee.

\algdef{SE}[SUBALG]{Indent}{EndIndent}{}{\algorithmicend\ }%
\algtext*{Indent}
\algtext*{EndIndent}

\DeclareMathOperator*{\argmin}{arg\,min} % thin space, limits underneath in displays
\DeclareMathOperator*{\argmax}{arg\,max} % thin space, limits underneath in displays

\begin{document}
% \linenumbers
\pagestyle{headings}
\mainmatter

\title{Automatic Rank Selection for High-Speed Convolutional Neural Network} % Replace with your title

\titlerunning{Automatic Rank Selection for High-Speed Convolutional Neural Network}

\authorrunning{Hyeji Kim, and Chong-Min Kyung}

\author{Hyeji Kim, and Chong-Min Kyung}

%Please write out author names in full in the paper, i.e. full given and family names. 
%If any authors have names that can be parsed into FirstName LastName in multiple ways, please include the correct parsing, in a comment to the volume editors:
%\index{Lastnames, Firstnames}
%(Do not uncomment it, because you may introduce extra index items if you do that...)

\institute{Korea Advanced Institute of Science and Technology\\
	Daejeon, Republic of Korea\\
	\email{hyejikim89@kaist.ac.kr}
}

\maketitle

\begin{abstract}
Low-rank decomposition plays a central role in accelerating convolutional neural network (CNN), and the rank of decomposed kernel-tensor is a key parameter that determines the complexity and accuracy of a neural network.
In this paper, we define rank selection as a combinatorial optimization problem and propose a methodology to minimize network complexity while maintaining the desired accuracy. 
Combinatorial optimization is not feasible due to search space limitations. 
To restrict the search space and obtain the optimal rank, we define the space constraint parameters with a boundary condition.
We also propose a linearly-approximated accuracy function to predict the fine-tuned accuracy of the optimized CNN model during the cost reduction. 
Experimental results on AlexNet and VGG-16 show that the proposed rank selection algorithm satisfies the accuracy constraint.
Our method combined with truncated-SVD outperforms state-of-the-art methods in terms of inference and training time at almost the same accuracy. 
\end{abstract}

\section{Introduction}
Convolutional Neural Networks (CNN) have been applied to the speech and vision-related tasks and have shown impressive performance. To achieve higher accuracy more complex deep neural network architectures are required, resulting in more computational power and memory. Such demands on compute power and memory makes it difficult to deploy CNNs to resource-constrained systems such as mobile and embedded devices.

There have been many parameter optimization techniques for reducing memory usage and accelerating CNNs including low-rank decomposition \cite{denil2013predicting,jaderberg2014speeding,kim2015compression,denton2014exploiting,zhang2016accelerating}, channel pruning\cite{He_2017_ICCV}, parameter pruning \cite{han2015deep,han2015learning} and quantization \cite{gong2014compressing,courbariauxbinarynet}. 
While pruning and quantization techniques can significantly reduce the parameter size and memory usage, it is hard to improve the practical inference runtime on commercial deep-learning platforms. They require special implementations such as sparse computation and low-precision libraries. To directly accelerate the inference work, methods in  \cite{denil2013predicting,jaderberg2014speeding,kim2015compression,denton2014exploiting,zhang2016accelerating} exploit low-rank decomposition where the convolutional and fully-connected layers are split into the low-complexity layers. X. Zhang \textit{et al.} \cite{zhang2016accelerating} and Y. He \textit{et al.} \cite{He_2017_ICCV} combine their parameter optimization techniques with Jaderberg \textit{et al.}'s \cite{jaderberg2014speeding} low-rank decomposition approach to further improve the accuracy.
%, since the rank of deeply decomposed layer parameter can be possible to precisely control.   
%Also, to compensate the accuracy drop, the low-rank decomposition has been applied 

In low-rank decomposition, the rank is the key parameter that determines the complexity of each layer. In other words, it is directly related to the memory usage, runtime and accuracy. Also, as discussed in \cite{zhang2016accelerating}, rank selection has critical effect on the classification accuracy to carefully select the rank of all layers.
% , some recent studies have been presented in \cite{kim2015compression,zhang2016accelerating}. 
Y-D Kim \textit{et al.} \cite{kim2015compression} exploit the global analytic solution of Variational Bayesian Matrix Factorization (VBMF) \cite{nakajima2013global} as a toolchain to determine the rank of each layer, and 
X. Zhang \textit{et al.} \cite{zhang2016accelerating} propose a layer-wise greedy strategy to determine the ranks of all layers satisfying the target complexity. 
This strategy assumes that the classification accuracy is roughly linear to the Principal Component Analysis (PCA) energy and defines an objective function to maximize the accumulated energy subject to the time complexity constraint.

% Figure1 ------------------------------
\begin{figure}[t]
\begin{center}
\subfigure[layer-wise search]{\includegraphics[width=0.46\linewidth]{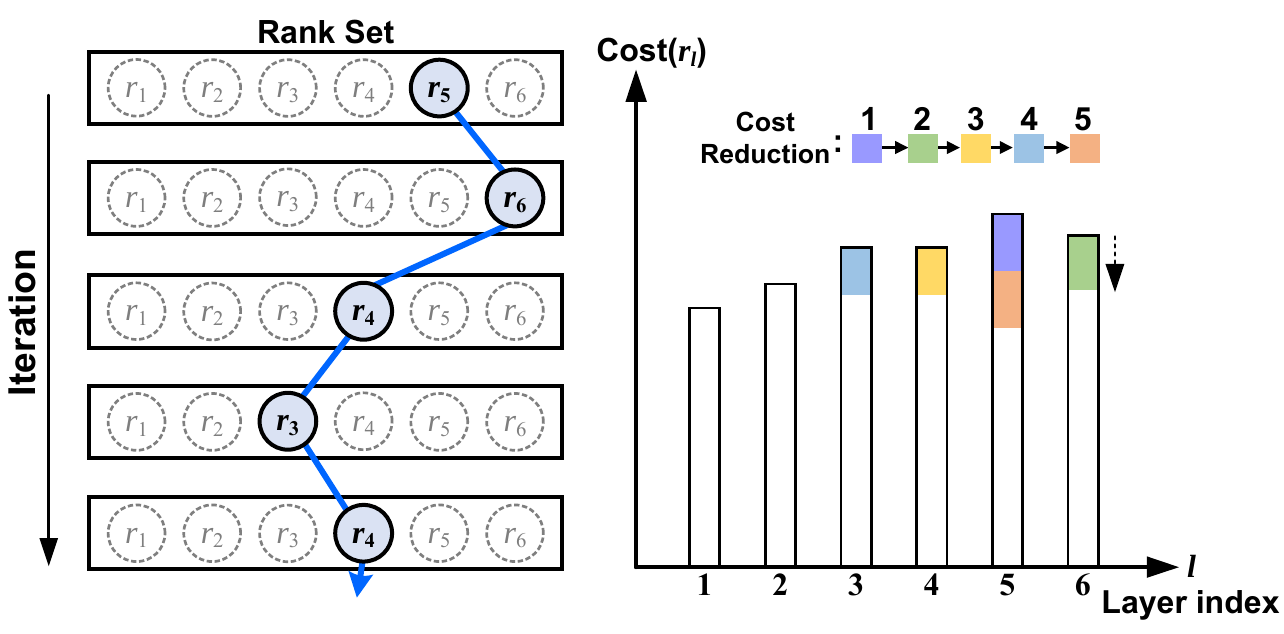}}
\hspace{1.0ex}
\subfigure[model-wise search]{\includegraphics[width=0.48\linewidth]{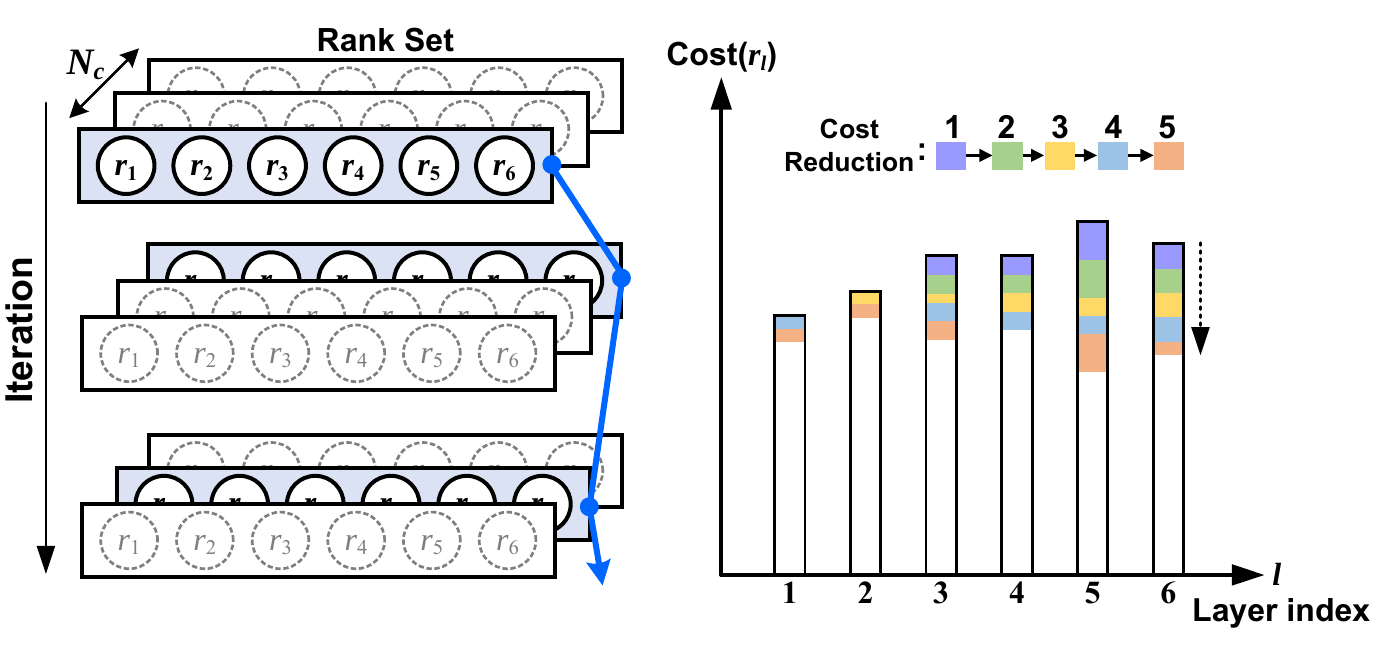}}
\end{center}
\vspace{-2.0ex}
   \caption{Rank selection strategy (with six kernel layers in a neural network as an example). (a) Legacy layer-wise greedy search selects a layer to reduce the rank for each iteration. (b) Proposed model-wise search selects a rank set to reduce the rank of all layers for each iteration. $N_c$ is the number of candidate rank sets.}   
\label{fig:fig1}
\vspace{-3.0ex}
\end{figure}

While these approaches can find the rank for each layer, 
the layer-wise greedy search \cite{zhang2016accelerating} is likely to be stuck in a local optima. 
% As shown in Fig.~\ref{fig:fig1}(a), the layer-wise greedy search selects a layer maximizing the whole-layer performance for each iteration, and reduces the rank of selected layer to compress the network complexity. For a decision, the performance is considered by modifying one of whole-layer rank at a time. 
As shown in Fig.~\ref{fig:fig1}(a), to reduce the network complexity (e.g., the number of operations or parameters) the layer-wise greedy algorithm reduces the rank of a layer at a time and selects a best layer maximizing the performance of network model over each iteration. 
% In this strategy, 
This layer-wise greedy algorithm always does not chooses an immediate worst-affected layer, so that it can prevent to find the best overall solution later.
% though the worse-layer selection can make better optimal solution in the future, CANNOT UNDERSTAND}. 
% though the selection of a current worst-affected layer can make better optimal solution in the future, 
% is not chosen, though the final optimal solution 
% a layer incurring the worst impact on the current model performance is not chosen, though the make near optimal solution in the future

On the other hand, the proposed model-wise greedy algorithm illustrated in Fig.~\ref{fig:fig1}(b) changes the rank of all layers at a time and iteratively selects a best set of rank for all kernel layers\footnote{In this paper, the word "kernel layer" refers to both the convolutional and fully-connected layers to be computed through the weight parameters.} (i.e. rank set) maximizing the performance of the network. % model in an iterative fashion. 
% 수정 필요
In this strategy, the performances of some candidate rank sets are compared to select a rank set including the immediate worst-affected layer. This allows to find a relatively optimal solution, since the candidate rank sets %in a iteration step %after a few steps 
are composed of various combinations of the rank. 
%regardless of whether the worst-affected layer is included or not.
%, including the cases with the worst-affected layer.
% since the effect of whole-layer rank is comprehensively evaluated.
However, the search space will significantly increase, unless the range of the ranks to be search is restricted.
For practical implementations, the candidate rank sets have to remain within a reasonable number as an exhaustive search over all rank sets will be computationally out-of-bounds.
% we have to the definition for effective search space with candidate rank sets is required, since the number of candidate rank sets is related to the total computation time of search algorithm. 
% and it is directly related to the computation time of search algorithm. 
% 왜 space 가 늘어나는지? : model-wise search algorithm은 
% For the practical implementation, 
% 현실적인 연산시간을 위해선, 모든 search space의 candidate rank set 을 통해 rank search algorithm을 수행할 수 없다. search space가 크고, 전체 모 집합 중, 특정 iteration 에서 random 하게 선택하여 compare 하는 rank set 의 표본 수가 너무 적다면, optimal solution에 가까운 candidate rank set을 선택 할 확률도 그만큼 줄어 든다. (refer 달면 좋음)

% To address the huge search space issue,
To address the above issues, we define rank selection as a combinatorial optimization problem and propose a method for model-wise rank selection with a space rejection rule. 
In our rank selection algorithm, we search a rank set achieving target accuracy and minimizing the computational cost. 
% overall algorithm -------
We define an accuracy function to estimate the desired final accuracy during cost reduction without fine-tuning.
We empirically observe that the test accuracy before fine-tuning (i.e. after low-rank decomposition) is almost linear to the recovered accuracy after several training epochs.
From this observation, the accuracy thresholds are determined and used as a termination condition of optimization algorithm and space rejection constraint. 
% From this observation, we predict the final recovery accuracy of optimized CNN model before fine-tuning, and we set the accuracy threshold as a space rejection constraint. 
Also, we define a cost function, which includes the rank of each layer 
and then generate the candidate rank sets that make almost same amount of reduction cost.
% Since the cost function is indeterminate equation, a huge amount of the solution vectors (rank set) can exist. 
Since the cost function is linear and multidimensional, a huge amount of solution vectors (i.e. rank set) can exist. 
To effectively find the optimal solution from the search space, we define space rejection parameters. 
This process is repeated iteratively until the target accuracy is achieved. 

% % The proposed methodology is illustrated in Fig~\ref{fig:fig2}.
% % In the proposed algorithm, we define the cost function depending on the rank of each layer, 
% We define a cost function, which includes the rank of each layer 
% and then generate the candidate rank sets that make almost same amount of desired reduction cost.
% % Since the cost function is indeterminate equation, a huge amount of the solution vectors (rank set) can exist. 
% Since the cost function is linear and multidimensional, a huge amount of solution vectors (i.e. rank set) can exist. 
% To find the optimal solution from the search space, we define space rejection parameters. 

% ----------
% step reduce restrict that the network complexity calculated from the candidate rank sets at a iteration step has almost same value which is dewithin 0.1\% error. Also 
% As a constraint of space rejection, the accuracy 
 
% space rejection 을 위한 constraint measure 로 test accuracy를 사용하였다. accuracy threshold function 을 정의 하기 위해, fine-tunign 전 (decomposition 후) test accuracy 와 fine-tuning 후 test accuracy 와의 관계를 선형 함수로 approximation 했다. 따라서 우리는  ...

In summary, our main contributions are as follows. 
(1) We propose a model-wise rank selection algorithm constrained on the desired accuracy. 
(2) We define the linear function of test accuracy before and after fine-tuning.
% (2) We show the linearity of accuracy before and after fine-tuning.
(3) The constraint parameters for the effective search space are defined to obtain an optimal solution. 
(4) We validate the performance of the proposed algorithm with state-of-the-art on the popular models such as VGG-16 and AlexNet. 
We show that the proposed optimal rank selection with Jaderberg \textit{et al.}, decomposition method \cite{jaderberg2014speeding} can provide outstanding performance when compared to recent studies \cite{kim2015compression,denton2014exploiting,zhang2016accelerating,He_2017_ICCV} for accelerating deep networks.
Furthermore, we expect the proposed rank selection algorithm can be effectively applied with not only the basic SVD but also state-of-the-art decomposition algorithms ~\cite{zhang2016accelerating,He_2017_ICCV}.

\section{Low-Rank Decomposition for DNN}

% Figure3 ------------------------------
\begin{figure} [t]
\begin{center}
\subfigure[spatial dimension]{\includegraphics[width=0.45\linewidth]{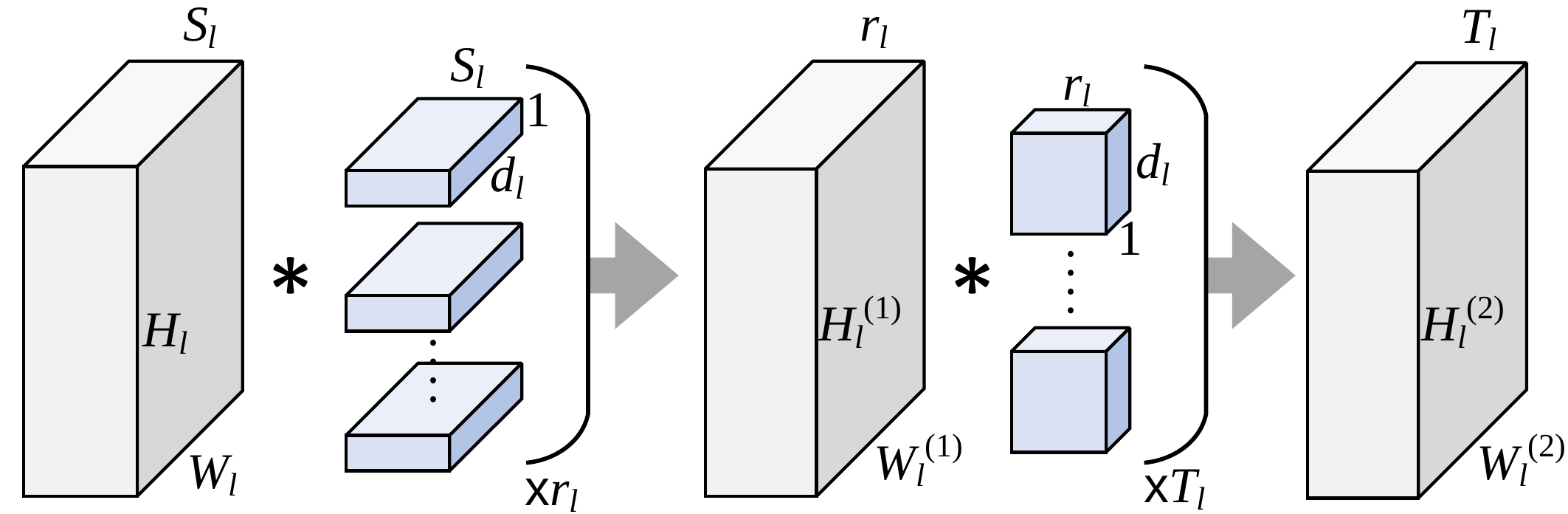}} 
\hspace{3.0ex}
\subfigure[channel dimension]{\includegraphics[width=0.45\linewidth]{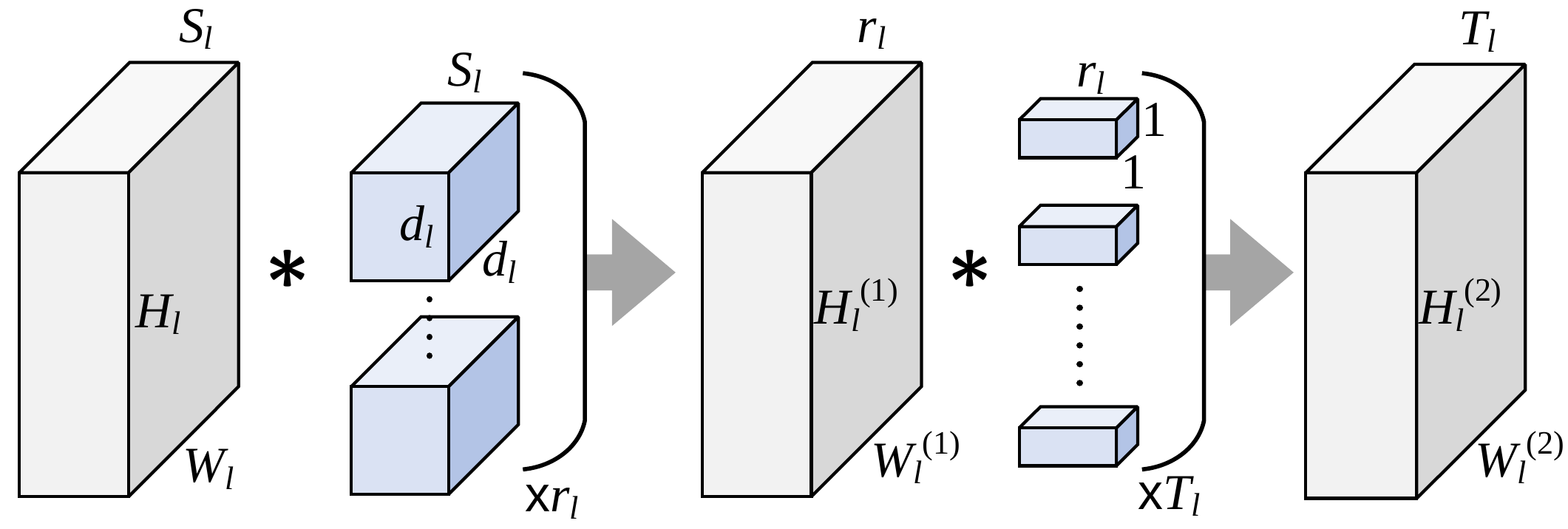}} 
\end{center}
\vspace{-2.0ex}
   \caption{2-level low-rank decomposition on (a) spatial dimension with ($d_{l}\times 1$), ($1\times d_{l}$) filter window, (b) channel dimension with ($d_{l}\times d_{l}$), ($1\times1$) filter window}
\label{fig:fig3}
\vspace{-3.0ex}
\end{figure}

% DNN 의 conv layer/fc layer를 어떻게 decompose하는지 그림으로 설명 & 수식으로 설명
% Matrix reshape 과정에서 filter/data간의 의미에 대해 설명[그림2]
% % #of parameter / # of cost 수식 정의 & 설명
% 특정한 rank 값에 의해 cost 가 결정됨을 설명
%---------------------------------------------------------------------
%(dxd)(1x1) : Convolution layer 압축 과정
%1. tensor unfolding (4-way to 2-way) : W(T, S, d, d) → W'(T, S*d*d)
%2. low-rank approximation : W'(T, S*d*d) → U(R, S*d*d) , V(T, R)
%3. reshape to 4-way tensor : U → U'(R, S, d, d) , V → V'(T, R, 1, 1)
%(1xd)(dx1) : Convolution layer 압축 과정
%1. tensor unfolding (4-way to 3-way) : W(T, S, d, d) → W''(T, S) x (d x d)  
%2. low-rank approximation : W''(T, S) x (d x d) → U(R, S) x (d x 1) , V(T, R) x (1 x d)
%3. reshape to 4-way tensor : U → U'(R, S, d, 1) , V → V'(T, R, 1, d)
%---------------------------------------------------------------------
% 목적 : 
% rank set 의 정의를 더 명확히 (직관적으로) 하기 위함
% rank 로 인해 complexity 가 결정됨을 설명해야 함
% general decomposition : 기본적인 k-level decomposition. 여기서 

% 본 논문의 가정 : 
% total complexity 는 original 을 넘으면 안됨, 즉, maximum rank 값을 정의함
% Decomposition method 는 주로 Jaderberg \textit{et al.} \cite{} 방식을 사용하며, first convolutional layer는 Denil \textit{et al.} \cite{} 방식을 사용함. (rank dimension 을 늘리기 위해서)

In CNNs, the convolutional layer has a 4-dimensional kernel tensor $\mathcal{K}\in \mathbb{R}^{d\times d \times S \times T}$, where $d$ is the spatial filter window size, $S$ is the number of input channels and $T$ is the number of output channels. 
By using low-rank decomposition, the 4-dimensional kernel tensor $\mathcal{K}$ can be decomposed into the matrix-multiplication of several small tensors with low-rank subspaces. 

We choose truncated singular value decomposition (SVD) for low-rank decomposition such that the tensor $\mathcal{K}$ of the $l$-th convolutional layer can be reshaped to the matrix $K\in \mathbb{R}^{d_{l}S_{l} \times d_{l}T_{l}}$. 
In a low-rank subspace, the matrix $K$ can be decomposed into 
$K^{(1)}K^{(2)}$
, where $K^{(1)} \in \mathbb{R}^{d_{l}S_{l} \times r_{l}}$ and $K^{(2)} \in \mathbb{R}^{r_{l} \times d_{l}T_{l}}$ with rank-$r_{l}$ \cite{jaderberg2014speeding}.

Fig~\ref{fig:fig3}(a) illustrates two decomposed kernel tensors for the $l$-th convolutional layer. 
Two matrices 
, $K^{(1)}$ and $K^{(2)}$,
are reshaped to 4-dimensional tensors with separate $d_{l}\times 1$ and $1\times d_{l}$ filter windows.
The rank of the $l$-th layer $r_{l}$ can be implemented as the number of filters whose sizes are $d_{l}\times 1 \times S_{l}$. 
Fig.~\ref{fig:fig3}(b) shows the case of separate $d_{l}\times d_{l}$ and $1\times 1$ filter windows \cite{denil2013predicting}.
In this case, matrix $K$ $\in$ $\mathbb{R}^{d^{2}_{l}S_{l} \times T_{l}}$ and the decomposed matrices $K^{(1)}$ $\in$ $\mathbb{R}^{d^{2}_{l}S_{l} \times r_{l}}$ and $K^{(2)}$ $\in$ $\mathbb{R}^{r_{l} \times T_{l}}$.

The original convolutional layer requires $d_{l}^{2}S_{l}T_{l}$ parameters and $d_{l}^{2}S_{l}T_{l}H_{l}W_{l}$ operations.   
From the spatial decomposition of $l$-th convolutional layer in Fig.~\ref{fig:fig3}(a), the number of parameters $p_{l}$ and operations $c_{l}$ are %can be defined as
% Eq1 : Number of decomposed parameter --------------------------
\begin{equation} \label{eqn:param}
p_{l}=
    d_{l}r_{l}(S_{l}+T_{l})
% \vspace{-2.5ex}
\end{equation}
% Eq4 : Number of computation (flops) for 2-layer decomposition --------------------------
\begin{equation} \label{eqn:oper}
    c_{l} =
    r_{l}d_{l}(S_{l}W_{l}^{(1)}H_{l}^{(1)}+T_{l}W_{l}^{(2)}H_{l}^{(2)})
.
\end{equation}
To decompose a fully-connected layer, the shape of filter window $d_{l}$ is fixed to 1.

In this paper, we restrict the number of parameters in the decomposed kernel tensors to less than or equal to the original 4-dimensional tensor.
Under this restriction, the maximum rank of each layer is given by
% Eq5 : r_max-initial --------------------------
\begin{equation} \label{eqn:rmaxinit}
    r_{l}^{max} =
    \lfloor S_{l}T_{l}d_{l}/(S_{l}+T_{l}) \rfloor.
\end{equation}

In Eq. (\ref{eqn:param}), the number of parameters is only controlled by the rank $r_{l}$ as $S_{l}$, $T_{l}$ and $d_{l}$ are constant values in a CNN model. 
Therefore, the rank of each layer can be adjusted to reduce the complexity of neural network.% such as $p_l$ and $c_l$ in Eq. (\ref{eqn:param}, \ref{eqn:oper}). 

\section{Approach : Rank Selection}

\begin{figure} [t]
\begin{center}
\subfigure[Rank Selection]{\includegraphics[width=0.67\linewidth]{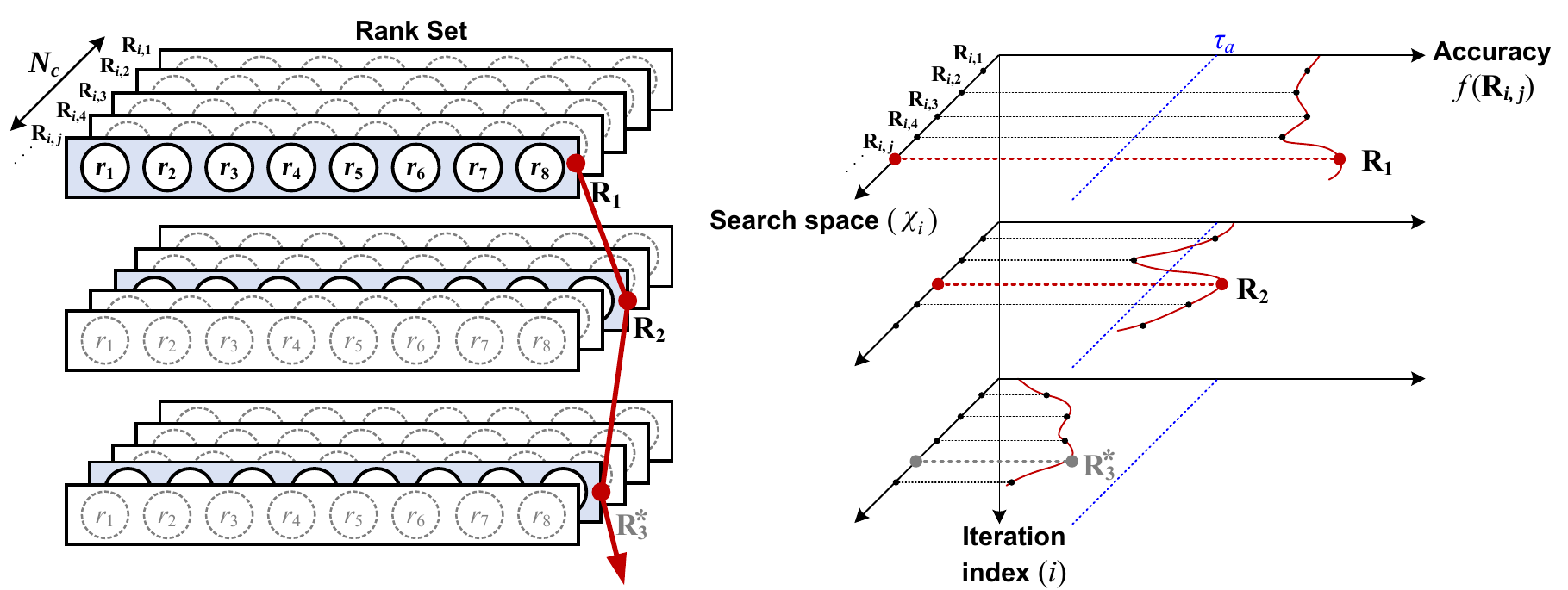}} 
\subfigure[Cost Reduction]{\includegraphics[width=0.32\linewidth]{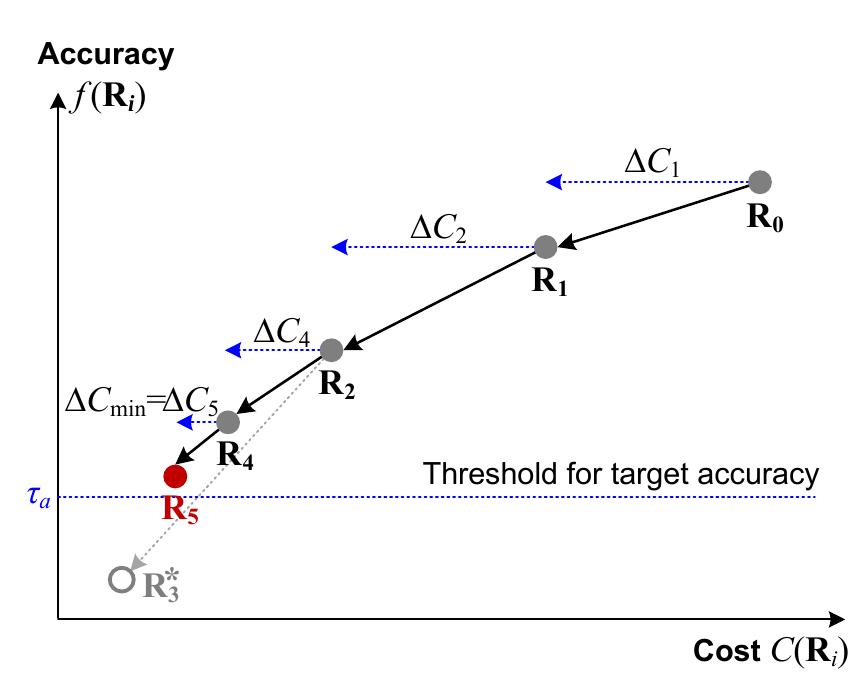}}
% \subfigure[]{\includegraphics[width=0.3\linewidth]{fig8_3_a.pdf}} 
% \subfigure[]{\includegraphics[width=0.37\linewidth]{fig8_3_b.pdf}}
% \subfigure[]{\includegraphics[width=0.315\linewidth]{fig8_3_c.pdf}}
% \includegraphics[width=0.96\linewidth]{graph1_6.pdf}
\end{center}
\vspace{-4.0ex}
\caption{Two types of view for iterative rank selection. Every iteration, (a) a rank set maximizing accuracy is selected among $N_c$ candidate sets, and (b) the cost is reduced until the accuracy is close to $\tau_a$}
% \caption{Process of iterative rank selection. (a)-(c) show three types of selection view. For each iteration, (a) select a rank set among $N_c$ candidate rank sets (b) maximizing the accuracy (c) by reducing the cost until the accuracy is close to threshold $\tau_a$.}
\label{fig:fig8}
\vspace{-3.0ex}
\end{figure}

% 전체 network model 에서 각 layer 의 rank (r_l) 를 element 로 하는 하나의 rank set 을 정의할 수 있다. 
% The proposed algorithm selects a rank set among the global rank combinations during a iteration. 
% definition of rank set :   
The goal of proposed algorithm is to search a rank set $\textbf{R}$ minimizing the cost $C(\textbf{R})$ such that the accuracy $f(\textbf{R})$ is greater than a threshold $\tau_{a}$ as shown in Fig.~\ref{fig:fig8},
% The objective function is as follow:
% Eq6 : objective function 1: total - (minimize cost s.t target accuracy) --------------------------
\begin{equation}\label{eqn:obj}
\argmin_{\textbf{R}} C(\textbf{R}), \quad \text{s.t.} \; f(\textbf{R}) > \tau_{a}
,
\end{equation}
where a rank set consists of the rank of each layer as $\textbf{R} = \{r_{1}, r_{2}, ..., r_{l}\}$.

The rank selection algorithm repeatedly applies the following four steps: 1) generating the candidate rank sets to achieve a smaller cost, 2) reducing the rank of kernel layers by using the candidate rank sets, 3) scoring the accuracy of each rank set, and 4) selecting a rank set maximizing the accuracy and satisfying an accuracy threshold.
After each iteration, a selected rank set is updated as an initial condition.

\subsection{Cost Function}

% How much reduce the rank. When we stop the reduction process.\cite{}
% we reduce the step value of cost, so we generate the candidate rank sets that make almost same network complexity. 
The cost function is the linear combination of the rank set with a scalar coefficient, i.e., 
\begin{equation}\label{eqn:cost1}
C_{m}(\textbf{R})=\sum\limits_{l=1}^{L}p_{l}r_{l}, \;\; C_{r}(\textbf{R})=\sum\limits_{l=1}^{L}c_{l}r_{l}
,
% \vspace{-1ex}
\end{equation}
% \begin{equation}\label{eqn:cost2}
% C_{r}(\textbf{R}_{k})=\sum\limits_{l=1}^{L}c_{l}r_{l,k}
% ,
% \end{equation}
where the $l$ is the index of kernel layer, $L$ is the number of optimized kernel layers in a CNN model. 
% ne of above cost functions in Eq.\ref{eqn:cost1}, \ref{eqn:cost2} is applied on the optimization problem as $C(\textbf{R}_k)$ in Eq.\ref{eqn:obj}.
The number of parameters $p_{l}$ and the number of operations $c_{l}$ represent the memory usage and runtime for the rank set $\textbf{R}$, respectively. 
Depending on the optimization type such as memory reduction or reducing runtime, 
$C_m(\textbf{R})$ or $C_r(\textbf{R})$ is used as the cost function $C(\textbf{R})$ in Eq. (\ref{eqn:obj}).

% \begin{table} [t]
% \centering
% % \caption{Operations and Parameters in AlexNet. 'Conv' involves all convolutional layers (\textit{Conv1}-\textit{Conv5}) and 'FC' involves all fully-connected layers (\textit{FC6}-\textit{FC8}).}
% \caption{The number of operations (FLOPs) and parameters (Weights) in AlexNet. 'Conv' involves all 5 convolutional layers and 'FC' involves all 3 fully-connected layers.}
% \renewcommand{\tabcolsep}{0.4cm} % Default value: 1
% \begin{tabular}{@{}crrcc@{}}
% \toprule
% \multicolumn{1}{c}{\begin{tabular}[c]{@{}c@{}}AlexNet\vspace{\tabgab}\end{tabular}} & \multicolumn{1}{c}{\begin{tabular}[c]{@{}c@{}} FLOPs (M)\end{tabular}} & \multicolumn{1}{c}{\begin{tabular}[c]{@{}c@{}} Weights (M)\end{tabular}}   \\ \midrule         
% {Conv}  & {665.7 (91.9\%)}   & {2.3 (3.8\%)\vspace{\cellgab}}\\
% {FC\hspace{0.2cm}}    & {58.6 (8.1\%)}    & {58.6 (96.2\%)}      \\ \bottomrule
% \end{tabular}
% % \vspace{1.0ex}
% \label{table:conv_param}
% \vspace{-2.0ex}
% \end{table}

% parameter 가 memory usage 가 되는 이유, operation 이 runtime 이 되는 이유 설명해야 함.
We can also configure the number of optimized layers $L$ to simplify the rank selection algorithm.
For the runtime optimization, only convolutional layers are optimized in general \cite{jaderberg2014speeding,zhang2016accelerating,He_2017_ICCV}.
% As an example of AlexNet in Table~\ref{table:conv_param}, all convolutional layers account for 91.9\% of the total number of operations, 
For example, all 5 convolutional layers of AlexNet account for 91.9\% of the total number of operations. 
% due to the multiplication with feature map.
% because Eq. (\ref{eqn:oper}) is calculated from the multiplication of parameter and feature map.
% , where the values in table are estimated by Eq\ref{eqn:cost1} and Eq\ref{eqn:cost2} with maximum rank in Eq\ref{eqn:rmaxinit}.
Note that the 2 fully-connected layers of AlexNet are primarily optimized for the memory usage, as these account for 96.2\% of the total number of parameters. 
% to find the solution rank set.

\subsection{Accuracy Constraints}

% Figure5 (graph1) ------------------------------
\begin{figure*} [b]
\begin{center}
% \subfigure[]{\includegraphics[width=0.3\linewidth]{graph1_a.pdf}} 
% \subfigure[]{\includegraphics[width=0.3\linewidth]{graph1_b.pdf}}
% \subfigure[]{\includegraphics[width=0.3\linewidth]{graph1_c.pdf}}
\includegraphics[width=0.94\linewidth]{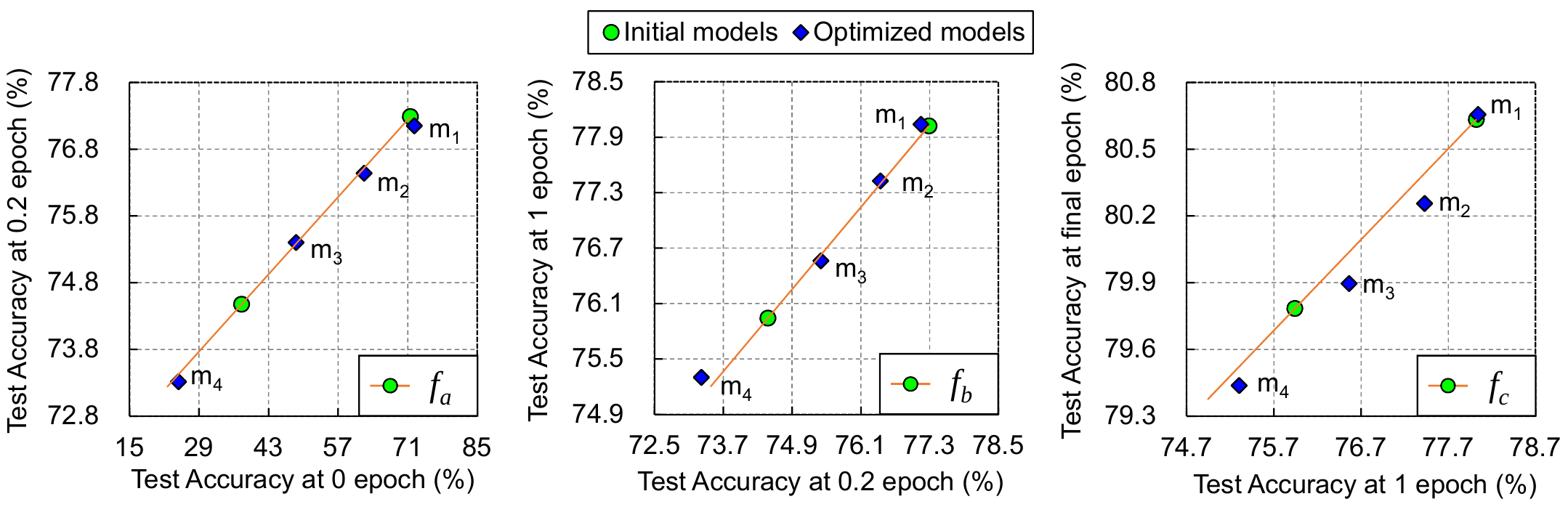}
\end{center}
\vspace{-2.0ex}
   \caption{AlexNet top-5 accuracy at 0, 0.2, 1, final training epochs up to 48 epochs (in our implementation). The blue points in graph are the measured accuracy of optimized CNN models having different cost. The green points are initial models to make the linear accuracy function before optimization. For each graph, the accuracy of different training epochs is almost linear} 
%    The accuracy function $(f_a, f_b, f_c)$ is linear regression of measured values}
\label{fig:graph1}
% \vspace{-3.0ex}
\end{figure*}

%for rank selection for each iteration
%to satisfy the target accuracy 
%compare the accuracy of 2 types network model

In order to confirm whether the selected rank set satisfies the target accuracy, a fine-tuning ($i.e$ training) stage is required.
Calculating the fine-tuned accuracy over every iteration will result in the selection algorithm taking too much time. % for fine-tuning. 
% To minimize the fine-tuning stage, %in the rank selection process,  
To this end, we propose approximating the accuracy function to roughly predict the fine-tuned accuracy during optimization. %in the optimization stage. 

We have experimentally observed that the test accuracy (top-5) before fine-tuning is almost linear to the fine-tuned test accuracy at the different training epochs, as shown in Fig.~\ref{fig:graph1}. 
We define each accuracy function as linear equation by:
\begin{equation}\label{eqn:fx}
% \begin{cases}
f_a = \alpha_ax + \beta_a, \;\, f_b = \alpha_bf_a + \beta_b, \;\, f_c = \alpha_cf_b + \beta_c
% f_a(x) = \alpha_ax + \beta_a,\;f_b(f_a) = \alpha_bf_a + \beta_b,\;f_c(f_b) = \alpha_cf_b + \beta_c
% \end{cases}
,
\end{equation}
where $x$ is the accuracy after optimization (0 epoch) and $f_a(x), f_b(f_a), f_c(f_b)$ are accuracy functions for different training epochs.
% 0.2, 1, final training epochs up to 48 epochs in our implementation.
% The final recovery accuracy $f_c$ is summarized with the coefficients in $f_a$,$f_b$ by:
% \begin{equation}\label{eqn:fc}
% \begin{cases}
% % {f_a} = \alpha_ax + \beta_a, \;\, f_b = \alpha_bf_a + \beta_b, \;\, f_c = \alpha_cf_b + \beta_c \\
% f_c (x)= \alpha x+ \beta \\
% \alpha = \alpha_c\alpha_b\alpha_a, \;\, \beta = \alpha_c\alpha_b\beta_a+\alpha_c\beta_b+\beta_c
% \end{cases}
% .
% \end{equation}
% To determine the coefficients in linear function, at least 2 optimized models are required. 

From the previous observation, we define three types of accuracy thresholds, $\tau_a, \tau_b, \tau_c$, to satisfy the target accuracy $\mu^*$ as
% Each threshold has the follow relationship:
\begin{equation}\label{eqn:tau}
% \begin{cases}
f_a(\tau_a) = \tau_b, \quad f_b(\tau_b) = \tau_c, \quad f_c(\tau_c) = \mu^*
% \end{cases}
.
\end{equation}
% , since $\tau_a$ is determined before fine-tuning.
If we define the accuracy function by training as many optimized models as possible up to the final epoch, 
% train the optimized models having different cost 
% use as many optimized models having different cost as possible and train them up to the final epochs, 
the accuracy function will be more accurate and it can precisely estimate the final accuracy 
% with only $\tau_a$. 
without $\tau_b$ and $\tau_c$.
However, %the training is quite time-consuming process.  
% due to the large 
to reduce training time and eliminate the latency during optimization due to the accuracy function, 
we use two initial models such as maximum-cost and half-cost models before the optimization process.
%, as shown in Fig.~\ref{fig:graph1}. 

Since the initial models are not from the rank optimization, 
the estimated accuracy from initial models can be different from the measured accuracy of optimized model.
Especially the accuracy at the 0 epoch (i.e. without fine-tuning) is directly affected by the optimization performance.
In other words, the threshold $\tau_a$ can have some uncertainty.
% Therefore, we check the fine-tuned accuracy up to maximum 1 training epoch to confirm that the last selected model satisfies target accuracy $\mu^*$.
Therefore, we use the accuracy thresholds $\tau_b$ and $\tau_c$ to confirm that the last selected model satisfies target accuracy $\mu^*$.
The threshold $\tau_c$ at 1 epoch is more reliable than $\tau_b$ at 0.2 epoch, since more training data is required to use $\tau_c$.
Nevertheless, we also use $\tau_b$ for early termination of the fine-tuning process before $\tau_c$ when the fine-tuned accuracy does not satisfy the threshold.

\subsection{Search Space Definition}

In our rank selection problem, the search space is defined by all possible combinations of elements in the vector spaces for rank.
% The total search space $\mathcal{X}$ is defined by the Cartesian product of vector spaces for rank, which are the sets involving the available rank element for each kernel layer.  
From the Cartesian product of vector spaces, the total search space $\mathcal{X}$ is defined as:
% ---- total search space : Cartesian product
% $\mathcal{X} =\prod\limits_{l=1}^{L}X_{l}=(x_1,...,x_L):x_l \in X_l \}$ 
% The total search space $\mathcal{X}$ is defined by the Cartesian product of vector space $X_l$ : 
\begin{equation}\label{eqn:m_X}
\mathcal{X} =\prod\limits_{l=1}^{L}X_{l}=\{\{r_1,...,r_L\}=\textbf{R} :r_l \in X_l \}%=\{\textbf{R}\}
% \mathcal{X} =\prod\limits_{l=1}^{L}X_{l}=\{(x_1,...,x_L)=\textbf{R} :x_l \in X_l \}%=\{\textbf{R}\}
% \{X_{l}\}_{l=1}^{L} \in \mathbb{R}^{L\times N^{(x)}}$ 
,
\end{equation}
where $r_l$ is the element of vector space $X_l$ for rank in $l$-th kernel layer and $r_l \geq 0$ for ${\forall}r_l\in\mathbb{Z}$. 
Each subset of $\mathcal{X}$ is a rank set %$\{r_1,...,r_L\}$, denoting a rank set  % $\textbf{R}_k$.
% ---- Rank set 
$\textbf{R} =\{r_{l}\}_{l=1}^{L}$.
% $\textbf{R}_{k} =\{r_{l}^{k}\}_{l=1}^{L} : r_{l}^{k} \in X_{l}$.

Space constraints are important to derive the optimal solution, since the search space without any constraints is an infinite field.
In this section, we define the parameters for appropriate search space with : 
1) upper and lower boundaries of rank, 2) step interval size of elements in vector space, 3) cost variance for iterative cost reduction, 4) cost margin to limit the amount of candidate rank sets. 
% ---------------
% We define the space constraints with following parameters: 
% 1) upper and lower boundary of rank, 2) step size of vector space for each rank, 3) cost variance for iterative cost reduction, 4) cost margin to limit the amount of rank set group. 
% ---------------
% 1) maximum rank $r_{l,i}^{max}$ and minimum rank $r_{l,i}^{min}$ for each layer, 2) step size $s_l$ of vector space for $l$-th layer rank, 3) cost variance $\Delta C_{i}$ for iterative cost reduction, 4) cost margin $\sigma_i$ to limit the amount of rank set group.
% 1) maximum rank $r_{l,k}^{max}$ and minimum rank $r_{l,k}^{min}$, 2) step size $s_l$ of vector space for rank, 3) cost variance $\Delta C_{k}$, 4) cost margin $\sigma_k$. 

\subsubsection{Boundary Condition}

\begin{figure}[t]
\begin{center}
\subfigure[Vector space $X_l$ of $r_l$]{\includegraphics[width=0.48\linewidth]{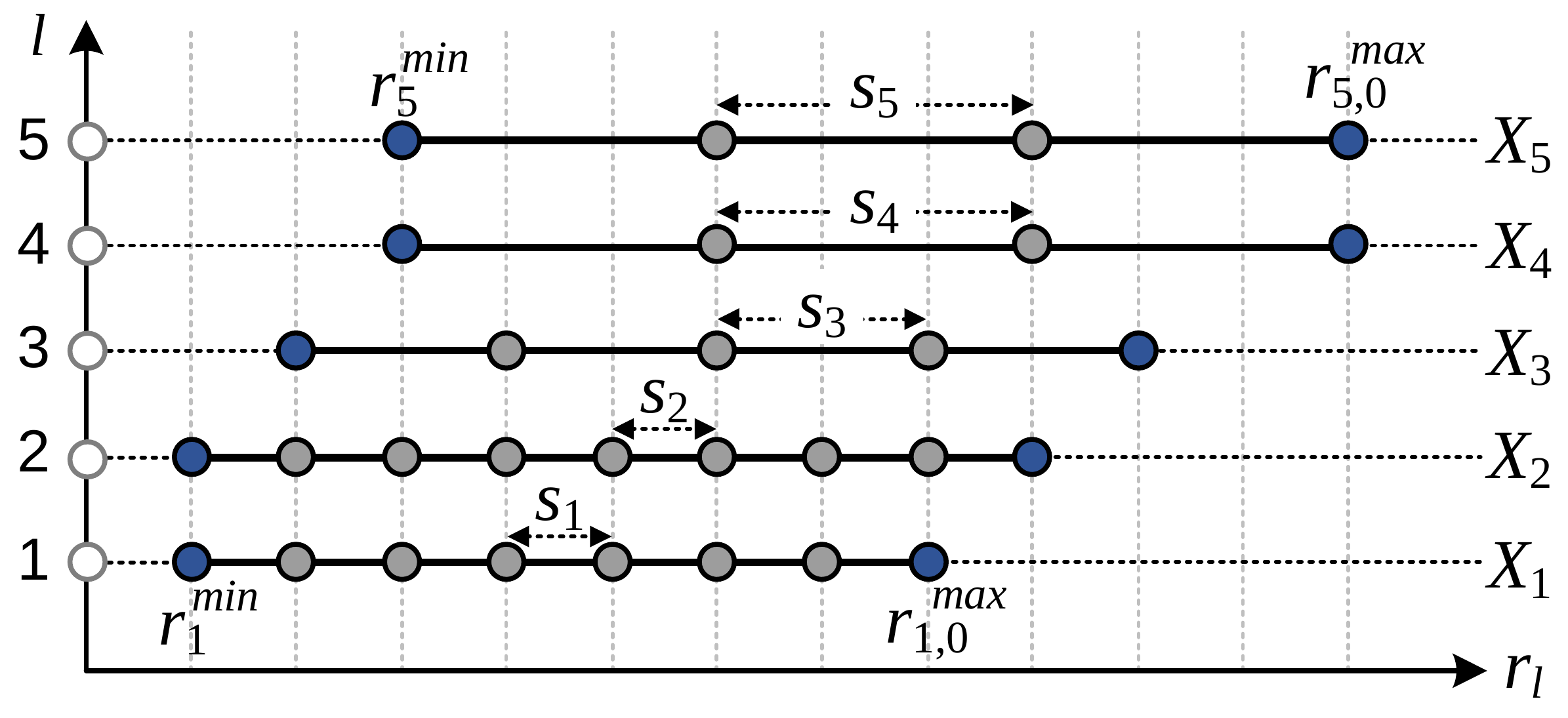}}
\hspace{2.0ex}
\subfigure[Vector space $\hat{X}_l$ of $\hat{r}_l$]{\includegraphics[width=0.48\linewidth]{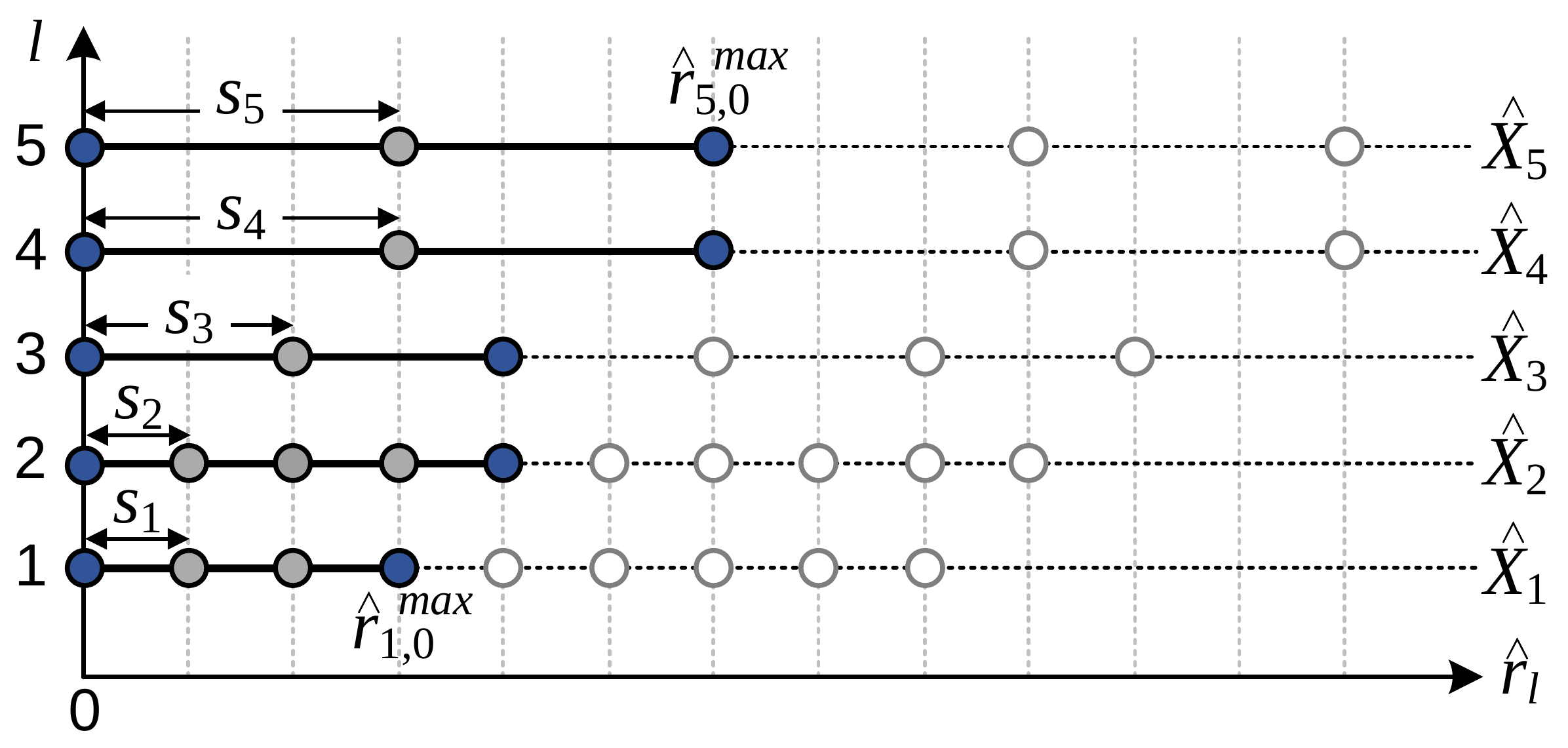}}
\end{center}
\vspace{-3.0ex}
   \caption{Vector spaces $X_l$ and $\hat{X}_l$ of rank in $l$-th kernel layer (show 5 layers as an example). The gray and blue dots are elements of each vector space. Respective elements in a vector space have the $s_l$ interval size. 
(a) Rank $r_l$ in $X_l$ is element of rank set $\textbf{R}=\{r_l\}_{l=1}^L$. (b) Rank $\hat{r}_l$ in $\hat{X}_l$ is element of candidate rank set $\Delta\textbf{R}=\{\hat{r}_l\}_{l=1}^L$}
% \label{fig:long}
% A rank set is the combination of elements in each vector spaces for rank
\label{fig:fig4}
\vspace{-2.0ex}
\end{figure}

As illustrated in Fig.~\ref{fig:fig4}(a), the vector space $X_l$ includes the available rank elements for each kernel layer.  
To restrict $X_l$, we set the upper boundary $r^{max}_{l}$ and lower boundary $r^{min}_{l}$, and the interval size $s_l$ of respective elements. 
The proposed model-wise search selects a rank set over each iteration, and a selected rank set $\textbf{R}_i=\{r_l^i\}_{l=1}^L$ is updated as the new maximum rank as 
% a selected rank set will be updated to the new upper boundary for each iteration. 
% Upper boundary : min(cost/rank, max rank)
\begin{equation}\label{eqn:max_r}
r^{max}_{l,i} =
\begin{cases}
r_{l}^{max} & \text{if{~}} i=0 \\ 
r_{l}^{i-1} & \text{otherwise}
\end{cases}
,
\end{equation}
where $i$ denotes the iteration index and $r_{l}^{max}$ is initial maximum rank in Eq. (\ref{eqn:rmaxinit}). 
The range of each $X_l$ gets smaller with every iteration. 
The interval size of elements in $X_l$ is $s_l = \max(1,\;\lfloor \delta_{s}r_l^{max} \rfloor )$.
Also, we define the minimum rank of each layer $r_l^{min}$ as a product of scaling factor $\delta_m$ as
% ---- step size : 1% of initial maximum rank
\begin{equation}\label{eqn:step}
% s_l = \max(1,\;\lfloor \delta_sr_l^{max} \rfloor ),\;\; r^{min}_{l} = \lceil \delta_m r_l^{max}\rceil.
r^{min}_{l} = \lceil \delta_m r_l^{max}\rceil.
% s_l = \max(1,\; r_l^{max}/\delta_s)
\end{equation}
% Also, we define the minimum rank of each layer $r_l^{min}$ as a product of scaling factor $\delta_m$ as 
% % lower boundary : 10% of initial maximum rank
% \begin{equation}\label{eqn:min_r}
% r^{min}_{l} = \lceil \delta_m r_l^{max}\rceil.
% \end{equation}
We empirically set the scaling factors, $\delta_s$ and $\delta_m$, to 0.01 and 0.1, respectively, 
thereby $\delta_s$ is 1\% of $r_l^{max}$ and $\delta_m$ is 10\% of $r_l^{max}$ in our implementation.

From the above results, we define the restricted vector space $X_{l}$ for rank by:
% ---- Elements in a vector space for l-th layer rank
\begin{equation}\label{eqn:vs}
X_{l} = \{r_l|\,r_l=n\,s_{l}, \;r_{l}^{min}\leq r_{l} \leq r_{l,i}^{max}\}
% X_{l} = \{r_l|r_l=n\cdot s_{l}, \;r_{l}^{min}\leq r_{l} \leq r_{l,i}^{max}\}_{n=0}^{N_{l}^x}
% $X_{l} = \{x_{n}|x_{n}=n\cdot s_{l}, \;r_{l}^{min}\leq x_{n} \leq r_{l}^{max}\}_{n=0}^{N_{l}^x}$
% $X_{l} = \{x_{l}^{n}|x_{l}^{n}=n\cdot s_{l}, \;r_{l}^{min}\leq x_{l}^{n} \leq r_{l}^{max}\}_{n=0}^{N_{l}^{(x)}}$
,
\end{equation}
where $n \in \mathbb{Z}$ and $r_l$ is the integer multiple of $s_l$ within the upper and lower boundary.
% Since the range of $X_l$ is smaller with every iteration due to the $r_{l,i}^{max}$ update, the search space will be also reduced.
% The restricted search space $\mathcal{X}^*$ is composed by the Cartesian product of $X_{l}^*$ and $\mathcal{X}^*\subset\mathcal{X}$.
% ---- the number of elements in a vector space for l-th layer rank.

% The complexity of $\mathcal{X}$ is $N^x =\prod_{l=1}^{L}N_{l}^x$. 
% The number of elements in $X_{l}$ is 
% $N_{l}^x = |X_{l}|=\lceil (r_{l}^{max}-r_{l}^{min})/s_{l}+1\rceil$.
% The complexity of $\mathcal{X}$ is computed by the product of $N_{l}^x$:
% % ---- the number of rank set combinations
% \begin{equation}\label{eqn:n_x}
% N^x =\prod\limits_{l=1}^{L}N_{l}^x.
% % N^x =\prod\limits_{l=1}^{L}N_{l}^x=\prod\limits_{l=1}^{L}\lceil (r_{l}^{max}-r_{l}^{min})/s_{l}+1\rceil.
% \end{equation}
% A space complexity of $N^x$ indicates that the number of rank sets is in $\mathcal{X}$.
% Since the range of $X_l$ is smaller with every iteration due to the $r_{l,i}^{max}$ update, the search space will be also reduced. 

% Figure4 ------------------------------

% \subsubsection{Sparse Subspace}
% Subsection text here.

\subsection{Candidate Rank Sets for Cost Reduction}
% define the available rank set (R_l) with step size(s_l) and maximum rank size(r_max,l)
% $N_{l}^{(x)} = \lceil\frac{r_{l}^{max}}{s_{l}}+1\rceil$
% how to select the candidate rank set group for a iteration.
% (we empirically set s = 2 in our experiments).

% The proposed model-wise greedy search in Fig.~\ref{fig:fig1} reduces the cost of CNN model by using a group of rank sets and select a rank set maximizing the accuracy.
% We reduce $C(\textbf{R})$ to select a rank set maximizing the accuracy.
% Every iteration, we reduce the cost of CNN model by $\Delta C_i=|C_{i}-C_{i+1}|$.
The proposed algorithm reduces the cost of CNN model by cost variance $\Delta C_i=|C_{i}-C_{i+1}|$ in every iteration.
The initial cost variance is defined by $\Delta C_0 = \lfloor \delta_{r}\,C_{max} \rfloor$, where $\delta_{r}$ is the cost scaling factor to determine the amount of cost reduction and $C_{max}$ is the maximum cost of the CNN model by applying $\{r_l^{max}\}_{l=1}^L$ in Eq. (\ref{eqn:rmaxinit}).  

To reduce the rank, we generate the rank sets $\{\Delta \textbf{R}\}$ resulting in $\Delta C_{i} \pm \sigma$, where $\sigma$ is the cost margin.
Without $\sigma$, the solution rank set may not exist, since the cost function is a multi-dimensional equation and the rank is an integer under our space limitation.   
% , and generate the rank set that makes $\Delta C_{i} \pm \sigma$, where $\sigma$ is cost margin.
% As described in Eq.\ref{eqn:cost1}, \ref{eqn:cost2}, the rank is the only parameter to determine $C(\textbf{R})$. 
By applying $\sigma$, we can define the reduction space $\mathcal{\hat{X}}_i$ satisfying $\Delta C_{i}\pm \sigma$ as: 
% ---- solution space satisfying the cost variance
\begin{equation}\label{eqn:sx_range}
\mathcal{\hat{X}} =\{\Delta \textbf{R} \,|\, \Delta C_i - \sigma \leq C(\Delta \textbf{R}) \leq \Delta C_i + \sigma\}
% $\mathcal{X}_i =\{\textbf{R}_{k}|C_k - \sigma_k \leq C(\textbf{R}_k) \leq C_{k} + \sigma_k, \textbf{R}_{k} \in \mathcal{X}\}$
.
\end{equation}
We define the set $\Delta \textbf{R}$ in $\mathcal{\hat{X}}$ as the candidate rank set. 
Also, $\mathcal{\hat{X}}$ can be represented by the Cartesian product of the vector space $\hat{X}_l$ for a candidate rank set.
% solution space with vector space
% \begin{equation}\label{eqn:Xi}
% \mathcal{X}_i^* =\prod\limits_{l=1}^{L}X_{l}^*=\{\{r_1^*,...,r_L^*\} :r_l^* \in X_l^* \}%=\{\textbf{R}\}
% .
% \end{equation}
As illustrated in Fig.~\ref{fig:fig4}(b), $\hat{X}_l$ has a maximum rank $\hat{r}_l^{max}$ and the interval of elements is same as Fig.~\ref{fig:fig4}(a).
% The vector space for candidate rank set is defined by :
\begin{equation}\label{eqn:vs_star}
\hat{X}_{l} = \{\hat{r}_l|\,\hat{r}_l=n\,s_{l}, \;0\leq \hat{r}_{l} \leq \hat{r}_{l,i}^{max}\}
% X_{l} = \{r_l|r_l=n\cdot s_{l}, \;r_{l}^{min}\leq r_{l} \leq r_{l,i}^{max}\}_{n=0}^{N_{l}^x}
% $X_{l} = \{x_{n}|x_{n}=n\cdot s_{l}, \;r_{l}^{min}\leq x_{n} \leq r_{l}^{max}\}_{n=0}^{N_{l}^x}$
% $X_{l} = \{x_{l}^{n}|x_{l}^{n}=n\cdot s_{l}, \;r_{l}^{min}\leq x_{l}^{n} \leq r_{l}^{max}\}_{n=0}^{N_{l}^{(x)}}$
,
\end{equation}
% According to $\Delta C_{i}$ and $r_{l,i}^{max}$, $\hat{r}_{l,i}^{max}$ is updated by:
% Upper boundary : min(cost/rank, max rank)
\begin{equation}\label{eqn:max_r}
\hat{r}^{max}_{l,i} = 
\lfloor \min(\Delta C_{i}/\epsilon_l,\;\; 2\delta_{r}\,r_{l,i}^{max}, \;\;r_{l,i}^{max}-{r}_{l}^{min}) \rfloor 
,
\end{equation}
where $\epsilon_l$ is the coefficient of cost function in Eq. (\ref{eqn:obj}).
The type of target cost determines $\epsilon_l$ as one of $p_l$ or $c_l$. 

% From the greedy selection, we subtract the candidate rank sets $\Delta \textbf{R}$ from the current rank set $\textbf{R}_{i-1}$, which is a selected rank set in the previous iteration. 
We generate the reduced rank sets $\{\textbf{R} |\, \textbf{R} = \textbf{R}_{i-1}-\Delta \textbf{R}, \Delta \textbf{R} \in \mathcal{\hat{X}}\}$
, and choose $\textbf{R}^*=\argmax_\textbf{R} f(\textbf{R})$.
% , and select a rank set $\textbf{R}^*$ maximizing the accuracy $f(\textbf{R})$.
% and select a rank set $\textbf{R}_i$ maximizing the accuracy $f(\textbf{R})$ and satisfying over $\tau_a$.
We update $\textbf{R}^*$ as $\textbf{R}_i$ only for $f({\textbf{R}}^*) > \tau_a$.
Otherwise,
% When $f({\textbf{R}}^*)$ is less than or equal to the accuracy threshold  $\tau_a$,
$\textbf{R}_i$ is still $\textbf{R}_{i-1}$ and the cost variance is reduced by: 
\begin{equation}\label{eqn:cost_i}
\Delta C_{i+1}=
\begin{cases}
\lfloor \Delta C_i/2 \rfloor & \text{if{~}} f(\textbf{R}^*) \leq \tau_a \\ 
\Delta C_i & \text{otherwise}
\end{cases}
.
\end{equation}
In detail, we consider that the cost variance is too large to find the optimal solution, and reduce $\Delta C_i$ by half.

% % Figure5 ------------------------------
% \begin{figure}[t]
% \begin{center}
% \includegraphics[width=0.50\linewidth]{fig4_2_b.pdf}
% \end{center}
% \vspace{-2.0ex}
%    \caption{Vector space $\hat{X}_l$ for candidate rank set. The right blue dot is the maximum rank $\hat{r}_{l,i}^{max}$ in $\hat{X}_l$. The gray and blue dots are elements of each vector space $\hat{r}_{l} \in \hat{X}_l$}
% % \label{fig:long}
% % A rank set is the combination of elements in each vector spaces for rank
% \label{fig:fig5}
% \vspace{-2.0ex}
% \end{figure}

% ======================================================
\subsection{Rejection Space}

% Figure6 ------------------------------
\begin{figure}[t]
\begin{center}
\includegraphics[width=0.54\linewidth]{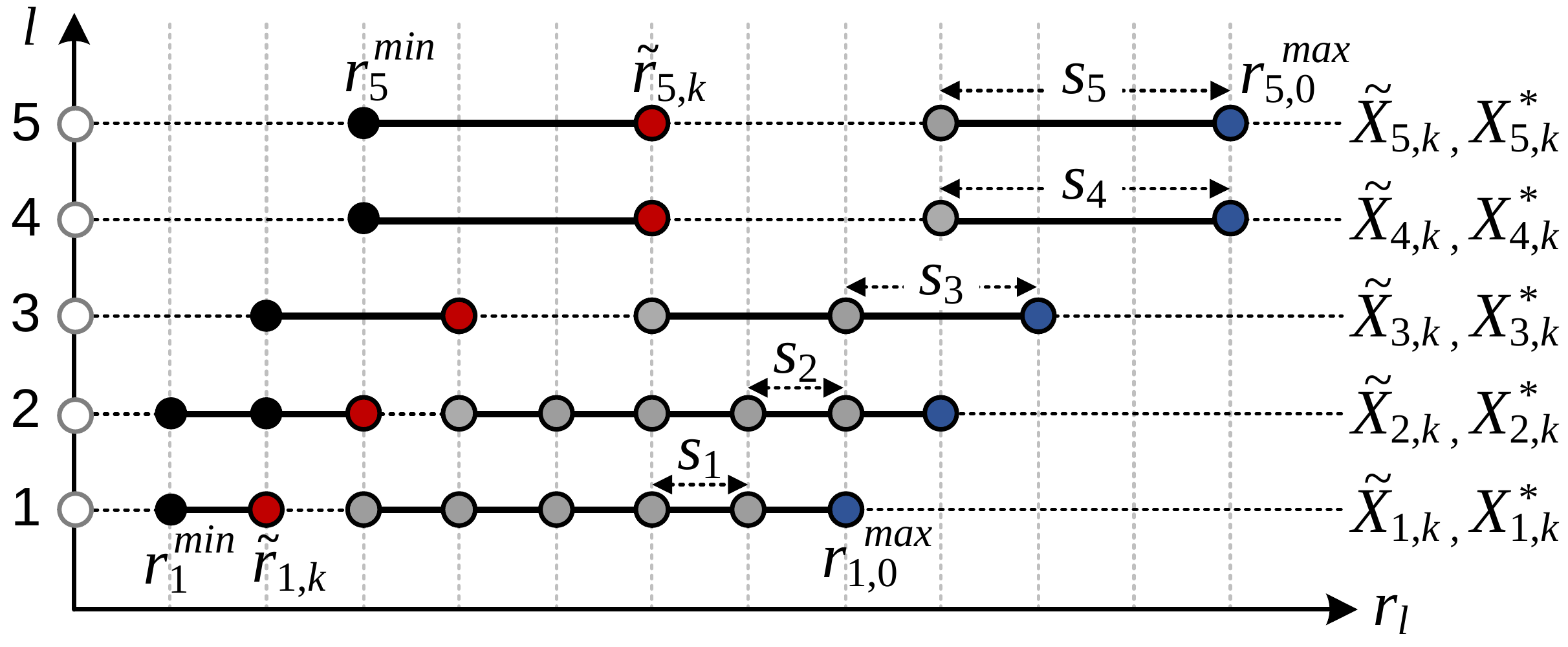}
\end{center}
\vspace{-3.0ex}
   \caption{Rejected and retained vector spaces, $\tilde{X}_{l.k}$ and ${X_{l,k}^{*}}$ respectively, for $k$-th rejection rank set $\tilde{R}_k$. The black and red dots are in $\tilde{X}_{l,k}$ and the red dot is the rejected rank $\tilde{r}_{l,k}$. 
The gray and blue dots in ${X}_{l,k}^*$ are the remain elements in ${X_l}$}
\label{fig:fig6}
\vspace{-2.0ex}
\end{figure}

% $\mathcal{Z}_i^*$
%remove unnecessary rank sets (at previous iteration, accuracy is under the threshold)
% To effectively find the optimal solution from the search space of rank set group, we define the space rejection parameters.
To further reduce the space complexity and retain valuable rank sets, we define the rejection space $\tilde{\mathcal{X}}$ and exclude $\tilde{\mathcal{X}}$ from $\mathcal{X}$. 
The strategy to maximize the probability of selecting the optimal rank set is to eliminate the unnecessary rank sets with the accuracy below $\tau_a$.
% rejection rank sets $\{\tilde{\textbf{R}}\}\subset \tilde{\mathcal{X}}$ that make the accuracy lower than $\tau_a$.

% tau 를 만족하지 못하는 rank set을 ~R로 정의. 
At every iteration, %we sort the candidate rank sets into the accuracy, then 
% the rank sets not satisfying $\tau_a$ are categorized as $\{\tilde{\textbf{R}}\}$.
% we can categorize the rank sets not satisfying $\tau_a$ as $\{\tilde{\textbf{R}}\}$.
we categorize the rejection rank sets $\{\tilde{\textbf{R}} |f(\tilde{\textbf{R}})\leq\tau_a\}$.% in the same cost region.
In addition to the observed rejection rank set, all smaller rank sets are included in the rejection space, since lower rank size corresponds to lower accuracy. 
As illustrated in Fig.~\ref{fig:fig6}, $X_l$ is separated into $\tilde{X}_{l,k}$ and ${X_{l,k}^{*}}$, which are rejected and retained vector spaces, respectively.
For $\tilde{\textbf{R}}_k \in \{\tilde{\textbf{R}}\}$, we define the vector space $\tilde{X}_{l,k}$ with the observed $\tilde{\textbf{R}}_k=\{\tilde{r}_{l,k}\}_{l=1}^L$ as the maximum rank by:
\begin{equation}\label{eqn:vs_x}
\tilde{X}_{l,k} = \{r_{l} |\,r_l=n\,s_{l}, \;r_{l}^{min} \leq r_{l} \leq \tilde{r}_{l,k}\}
,
\end{equation}
where $k$ is the index of rejection rank set.
% From the Cartesian product of $\{\tilde{X}_{l,k}\}_{l=1}^L$, we can define the rejection subspace $\tilde{\mathcal{X}}_k$ from $\tilde{\textbf{R}}_k$.
From the Cartesian product of rejection vector space $\tilde{X}_{l,k}$, we can define the rejection subspace $\tilde{\mathcal{X}}_k$. 
% $\tilde{\mathcal{X}}_k=\prod_{l=1}^{L}\tilde{X}_{l,k}$.
% from $\tilde{\textbf{R}}_k$.
The total rejection space $\tilde{\mathcal{X}}$ is the union of all $\tilde{\mathcal{X}}_k$, 
\begin{equation}\label{eqn:total_reject_x}
% \tilde{\mathcal{X}_k}=\prod\limits_{k=1}^{N_k}\tilde{X}_k,
% \tilde{\mathcal{X}} = \bigcup\limits_{k=1}^{N_k}\tilde{\mathcal{X}_k}=\{{\textbf{R}}|\;\text{for some }k\in\{ 1,...,N_k \}, {\textbf{R}}\in\tilde{\mathcal{X}}_k\}
% \tilde{\textbf{R}}\in\tilde{\mathcal{X}}_k\}
\tilde{\mathcal{X}}_k=\prod_{l=1}^{L}\tilde{X}_{l,k},\;\;\; \tilde{\mathcal{X}} = \bigcup\limits_{k=1}^{N_k}\tilde{\mathcal{X}_k},
\end{equation}
where $N_k$ is the number of rejection sets in the candidate rank sets.
Therefore, we retain the effective search space ${\mathcal{X}}^*$ as the complement of $\tilde{\mathcal{X}}$ in ${\mathcal{X}}$ by:
\begin{equation}\label{eqn:total_remain_x}
{\mathcal{X}}^* = {\mathcal{X}}\cap\tilde{\mathcal{X}}^{\complement}=\{ {\textbf{R}}:{\textbf{R}}\in{\mathcal{X}} \;|\; \textbf{R}\notin \tilde{\mathcal{X}} \}
.
\end{equation}
The remaining rank set in ${\mathcal{X}}^*$ will be considered for next iteration search.

\subsection{Algorithm Procedure}

\begin{algorithm}[t]
    \caption{Model-wise Automatic Rank Search}
    % \text{\small{(before fine-tuning)}}}
    \label{alg:code}
    \begin{algorithmic}[1] \small 
    \Statex \textbf{// Stage1 : Cost minimization subject to accuracy constraint $\tau_a$}
	\State Initialize the target cost $\epsilon_l$, optimization layer $L$, space constraints $\{r_l^{max}, r_l^{min}, s_l\}$ and reduction cost $\Delta C_0$
%     and space constraints $\{r_l^{max}, r_l^{min}, s_l\}$
%     \Statex space constraints $\{r_l^{max}, r_l^{min}, s_l\}$ and reduction cost $\Delta C_0$
% 	\State $i=0, j=0$ 
    \State \textbf{repeat}    	
    \Indent
        \State{Update the parameters $\{r^{max}_{l,i}, \hat{r}^{max}_{l,i}, \Delta C_i\}$ }
%         \State{Define the search space $(\mathcal{X}_i\cap\mathcal{X}_{i-1}^*)$}  
%         \State{Generate the reduction space $\hat{\mathcal{X}}_i$ satisfying $\Delta C_i$ }
%         \State{Extract $N_c$ candidate rank sets $\{\Delta \textbf{R}\}$ in ${\mathcal{X}} \cap \tilde{\mathcal{X}}^{\complement}$}
%         \State{Extract $N_c$ candidate rank sets $\{\Delta \textbf{R}\}$ excluding $\{\Delta \textbf{R}|\tilde{\textbf{R}}=\textbf{R}_{i-1}-\Delta \textbf{R}\}$ }
        \State{Extract $N_c$ candidate rank sets $\{\Delta \textbf{R} | (\textbf{R}_{i-1}-\Delta \textbf{R})  \notin \tilde{\mathcal{X}}_{i-1}\}  $ from reduction space $\hat{\mathcal{X}}_i$ }
%         \State{Extract $N_c$ candidate rank sets $\{\Delta \textbf{R}\}$ excluding $\{\textbf{R}_{i-1}-\Delta \textbf{R}\} \in \tilde{\mathcal{X}}_{i-1}$ }
    	\For{each rank set in $\{\Delta \textbf{R}\}$ } {}
%           \State{Reduce the cost from $\textbf{R} = \textbf{R}_{i-1}-\Delta\textbf{R}$ }
          \State{Check the accuracy $f(\textbf{R})$ from $\textbf{R} = \textbf{R}_{i-1}-\Delta\textbf{R}$}           
          \State{Group the rejection set $\{\tilde{\textbf{R}} |\tilde{\textbf{R}} = {\textbf{R}}, f(\tilde{\textbf{R}})\leq\tau_a\}$ }
      	\EndFor
        \State Define the rejection vector spaces $\tilde{X}_{l,k}$ from each rejection set $\tilde{\textbf{R}}_k \in \{\tilde{\textbf{R}}\}$
%         \State Generate the rejection space $\tilde{\mathcal{X}}_i = \tilde{\mathcal{X}}_{i-1}\cup_k \tilde{\mathcal{X}}_{k}$  
         \State Generate the rejection space $\tilde{\mathcal{X}}_i = \tilde{\mathcal{X}}_{i-1} \bigcup_k (\prod_l{\tilde{X}_{l,k}}) $  
%           \State Generate the rejection space $\tilde{\mathcal{X}}_i = (\prod_l{\tilde{X}_{l,k}}) \bigcup_k \tilde{\mathcal{X}}_{i-1}$  
        \State Select a rank set $\textbf{R}^*=\argmax_{\textbf{R}} f(\textbf{R})$ 
        \State Update the list of selected rank set $[\textbf{R}] \leftarrow \textbf{R}_i=\textbf{R}^*$ only for $f(\textbf{R}^*)>\tau_a $.
%         Otherwise, $\textbf{R}_i=\textbf{R}_{i-1}$. 
        \Statex \hspace{\algorithmicindent} Otherwise, $\textbf{R}_i=\textbf{R}_{i-1}$.
%         \State {$i \leftarrow i+1$}
        \State {$i = i+1$}
        \EndIndent    
%     \State \textbf{until} $\exists i, \text{such that} f(\textbf{R}_i) \leq \tau_a$        
%     \State \textbf{until} $f(\textbf{R}_i) > \tau_a$ and $\Delta C_i\ < \Delta C_{min}$       
    \State \textbf{until} $f(\textbf{R}^*) \leq \tau_a$ and $\Delta C_i\ \leq \Delta C_{min}$  
    \Statex \textbf{ }
    \Statex \textbf{// Stage2 : Fine-tuning and accuracy check with threshold $\tau_b$ and $\tau_c$}
%     for more precise target accuracy}
    \State {$\textbf{R}_j$ = last updated rank set in $[\textbf{R}]$}
    \State \textbf{repeat}
    \Indent
%     \State Fine-tune CNN model of $\textbf{R}_{i}$ and check $\tau_b, \tau_c$.
% 	\State {$\textbf{R}_j$ = last updated rank set in $[\textbf{R}]$}
	\State {Briefly fine-tune the network model using $\textbf{R}_j$ till 1 epoch}
%     \State {Check the test accuracy at the intermediate training epochs}
%     \Statex {\algorithmicindent} and check the accuracy 
	\State{\textbf{If} $f(\textbf{R}_j) < \tau_b$ or $f(\textbf{R}_j) < \tau_c$ }
    \Indent
    	\State $j = j+1$
    	\State{$\textbf{R}_{j}$ = previous updated rank set of $\textbf{R}_{j-1}$ in [\textbf{R}]}
        
%         \State $j \leftarrow j+1$
    \EndIndent
	\EndIndent    
    \State \textbf{until} $f(\textbf{R}_j) \ge \tau_b$ and $f(\textbf{R}_j) \ge \tau_c$
    \State Fine-tune the network model using the last selected $\textbf{R}_{j}$ for final epoch
	\State \textbf{Output : fine-tuned model using $\textbf{R}_{j}$} %{\small(fine-tuned)}.
    \end{algorithmic}
%    }
\end{algorithm}
% \vspace{-4.0ex}

The rank search procedure is formalized in Alg.~\ref{alg:code}.
The proposed algorithm aims to iteratively minimize the cost satisfying the target accuracy.
% ($i.e.$ memory usage or runtime) 
% by reducing the rank of kernel layers.
At first, the space constraints for boundary condition, $\{r_l^{max}$,$r_l^{min},s_l\}$, are initialized with the scaling factors, $\{\delta_s, \delta_m, \delta_r\}$, when the target cost and number of optimization layers are determined.
At every iteration, the search space is redefined by updating the constraint parameters and removing the unnecessary search space. 
The candidate rank sets $\Delta \textbf{R}$ are extracted from the search space.
For each candidate rank set, the accuracy of the reduced rank set is scored. 
Then, a rank set $\textbf{R}^*$ that provides maximum accuracy is selected, 
and $\textbf{R}_i$ is updated in the list $[\textbf{R}]$ if the accuracy is higher than threshold $\tau_a$.% as illustrated in Fig.~\ref{fig:fig8}.
This procedure repeats iteratively until the accuracy of $\textbf{R}^*$ and the reduction cost $\Delta C_i$ are lower than the $\tau_a$ and minimum cost reduction $\Delta C_{min}$, respectively. 

% For more precise targeting the desired accuracy, we briefly fine-tune a CNN model of the last selected rank set $\textbf{R}_i$ after the rank search is done.
After cost reduction, we fine-tune a CNN model of the last selected rank set in the list $[\textbf{R}]$ up to 1 epoch.% to achieve the desired accuracy.
The accuracy at 0.2 epoch and 1 epoch is verified whether it satisfies the threshold $\tau_b$ and $\tau_c$.
% We confirm whether the accuracy below 1 epoch is over the threshold $\tau_b$ and $\tau_c$. 
% Then, we check the accuracy of several training epochs to confirm whether the accuracy is over the threshold $\tau_b$ and $\tau_c$. 
When the accuracy condition is not satisfied, we fine-tune another network model with previous updated rank set in $[\textbf{R}]$ and check the accuracy again until all accuracy thresholds are satisfied. 
% The last selected set satisfies all accuracy thresholds.
% We repeat the short fine-tuning, until the accuracy condition is satisfied.
Finally, an optimized CNN model using the last selected rank set is fine-tuned for the final epoch.

% 이미 앞에서 다 한얘기들..!!!
% In the proposed algorithm, we generate the candidate rank set group that makes the almost same network complexity (e.g. the number of operations or parameters).
% The rank of each layer is reduced from the candidate rank set group, and the accuracy of modified network model is scored. Then, a rank set that makes maximum accuracy is taken.
% As a constraint of space rejection, we define the minimum accuracy threshold. 
% % The subspaces of the unnecessary rank set that does not satisfy the accuracy threshold will be excluded from the search space and the remaining subspaces are considered for next iteration search. 
% The subspaces of the unnecessary rank set will be excluded from the search space and the remaining subspaces are considered for next iteration search. 
% % We empirically observe that the test classification accuracy before fine-tuning (i.e. after low-rank decomposition) is roughly linear on the recovered accuracy at the several training epochs. 
% This process is iteratively repeated until the target complexity or accuracy is satisfied. 

% \subsubsection{Rank-Set Selection}
% Subsection text here.

% \subsubsection{Stopping Condition}
% stop when the accuracy condition is not satisfied
% there are 2 types of accuracy threshold, it have to satisfy that two condition
% accuracy before fine-tuning
% accuracy after fine-tuning at 1/10 epoch
% \vspace{-2.0ex}
\subsubsection{Optimization Time}

% (Hyeji) 휴리스틱 방식의 한계와, processing time 을 현실적으로 제한하기위한 기법으로 무엇을 썼는지 소개. 
% further 속도 향상을 위해 partial set 을 사용한 대신, Accuracy prediction based Space rejection rule 기법을 이용하여 성능저하를 막음.  
% In the proposed algorithm, 
There are three parts of the optimization process, which are the generation of candidate rank sets, accuracy check and brief fine-tuning. 
The most time consuming part is the accuracy check in the cost minimization step (i.e. Step1). 
The total time of accuracy check is proportional to the number of candidate sets. 
Therefore, in every iteration we only test the randomly extracted $N_c$ sets to maintain a reasonable amount of optimization time. 
% Since there can be the huge amount of candidate rank sets due to the combinatorial problem, 
% despite of due to the combinatorial problem 
% the number of rank set to verify the test accuracy determines the almost search time. 
% extracted $N_c$ sets among the large amount of combinations.
However, the smaller the number of extracted subsets, the lower the probability of obtaining a optimal solution. 
Therefore, we incrementally reject the rank sets that are predicted to have a lower accuracy by using the method of Sec. 3.2.
% from the proposed accuracy function in Sec. 3.2. 
This space rejection increases the probability of having a near optimal solution with the extracted subset.
Also, we set the intermediate accuracy threshold such as $\tau_b$ before $\tau_c$ for early termination of the fine-tuning stage.

\section{Experiments}

%%%%%%%%%%%%%%%%%%%%%%%%%%%%%%%%%%%%%%%%%%%%%%%%%%%%%%%%%%%%%%%%%%%%%%%%%%%%%%%%%%%%%%%%%%%%%

% (Hyeji) target accuracy 에 대해 cost minimized 알고리즘임을 강조

% We evaluate that our rank selection algorithm can appropriately find the cost minimized model satisfying the target accuracy. 
In our experiments, we use two CNN models: AlexNet \cite{NIPS2012_4824} and VGG-16 \cite{simonyan2014very}.
AlexNet has 5 convolutional layers and 3 fully-connected layers, and 
VGG-16 has 13 convolutional layers and 3 fully-connected layers.
The baseline top-5 accuracy of AlexNet is 80.03\% \cite{NIPS2012_4824} and VGG-16 is 89.9\% \cite{simonyan2014very} on the ImageNet 2012 validation set \cite{deng2009imagenet} of 1000 classes.
In the fine-tuning and evaluation stage, we crop 227x227 size images for AlexNet and 224x224 size images for VGG-16.
For fine-tuning VGG-16, the ImageNet dataset is scaled with a fixed smallest side 256 as described in \cite{simonyan2014very}.
We use Berkeley's \textit{Caffe} \cite{jia2014caffe} for the implementation.

% Before applying the proposed rank selection,
% Jaderberg \textit{et al.} method \cite{}. 
In the first step, we decompose the CNN model by using Denil \textit{et al.} \cite{denil2013predicting} and Jaderberg \textit{et al.} methods \cite{jaderberg2014speeding}. 
For the first convolutional layer, we adopt the 2-level channel decomposition \cite{denil2013predicting} to separate $d\times d$ kernel window into the $d\times d$ and $1\times 1$ windows.% (see Fig.~\ref{fig:fig3}(b)).
Since the input channel size of first layer is small due to the RGB image, the method in \cite{denil2013predicting} is better solution to increase the number of decomposed filters than the method in \cite{jaderberg2014speeding}. 
For other kernel layers, we adopt the 2-level spatial decomposition \cite{jaderberg2014speeding} to separate $d\times d$ kernel window into the $d\times 1$ and $1\times d$ windows. %(see Fig.~\ref{fig:fig3}(a)).

For the initialization of rank selection algorithm, 
we empirically set the minimum rank of each layer $r_l^{min}$ to 10\% of $r_l^{max}$ and the interval size $s_l$ of the vector space is 1\% of $r_l^{max}$ in our implementation. 
To further speedup the algorithm, we restrict the number of candidate rank sets $N_c$ to maximum 200. Also, we use 10\% of the validation dataset to check the accuracy for the selection procedure.
% We provide more details about our algorithm in the supplementary material.

\subsection{Experiments with AlexNet}

We define the the target accuracy and target cost from \cite{kim2015compression} using tensor decomposition on AlexNet.
The cost of AlexNet is minimized until the \textit{accuracy} is almost same as the balanced optimization approach \cite{kim2015compression}.

\subsubsection{Balanced optimization}
% Rank selection procedure
% Training policy
% Performance comparison 

\begin{table}[t]
\centering
\caption{Performance comparison for balanced optimization}
\vspace{-1.0ex}
\renewcommand{\tabcolsep}{0.1cm} % Default value: 1
\begin{tabular}{@{}cccccccc@{}}
\toprule
AlexNet         & \multicolumn{1}{c}{FLOPs}                             & \multicolumn{1}{c}{Weights}                                & \multicolumn{1}{c}{Top-5 Acc.} & \multicolumn{1}{c}{Target Acc.} & \multicolumn{1}{c}{Decomp.} & \multicolumn{1}{c}{Rank Sel.} \\ \midrule
Y-D Kim~\cite{kim2015compression} & \begin{tabular}[c]{@{}c@{}}272.0 M\vspace{\tabgab}\\ ($\times$2.67)\end{tabular} & 
\begin{tabular}[c]{@{}c@{}}11.0 M\vspace{\tabgab}\\ (-81.6 \%)\end{tabular} & 
{78.33\%}  & - & 
% 3-level Tucker & 
\begin{tabular}[c]{@{}c@{}}{Tucker}\vspace{\tabgab}\\ {(3-level)}\end{tabular}  & 
VBMF \cite{nakajima2013global}                          \vspace{\cellgab}\\
% \textbf{{Ours} (fine-tuned)}       & 
\begin{tabular}[c]{@{}c@{}}{\textbf{Ours}}\vspace{\tabgab}\\ {(final model)}\end{tabular} &
\begin{tabular}[c]{@{}c@{}}\textbf{238.5 M}\vspace{\tabgab}\\ \textbf{($\times$3.04)}\end{tabular} & 
\begin{tabular}[c]{@{}c@{}}10.6 M\vspace{\tabgab}\\ (-82.6 \%)\end{tabular} & 
{78.43\%} & {78.33\%\cite{kim2015compression}} & 
% 2-level SVD & 
\begin{tabular}[c]{@{}c@{}}{SVD}\vspace{\tabgab}\\ {(2-level)}\end{tabular}  & 
\begin{tabular}[c]{@{}c@{}}{Model-wise}\vspace{\tabgab}\\ {Rank Search}\end{tabular} & 
\hspace{0.5ex}                            \\ \bottomrule
\end{tabular}
\label{table:graph2_b}
\vspace{-3.0ex}
\end{table}

\begin{figure}[t]
\begin{center}
% \includegraphics[width=0.99\linewidth]{acc_fc.pdf}
% \subfigure[Cost Reduction]{\includegraphics[width=0.45\linewidth]{Fig7a.pdf}} \\
\subfigure[Iterative cost reduction]{\includegraphics[width=0.49\linewidth]{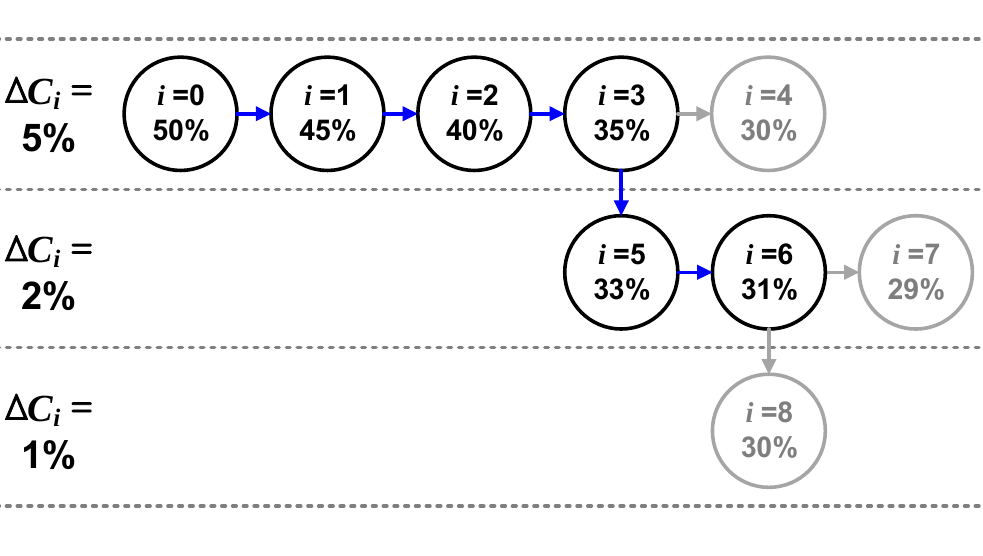}} 
\subfigure[Accuracy of optimized models]{\includegraphics[width=0.44\linewidth]{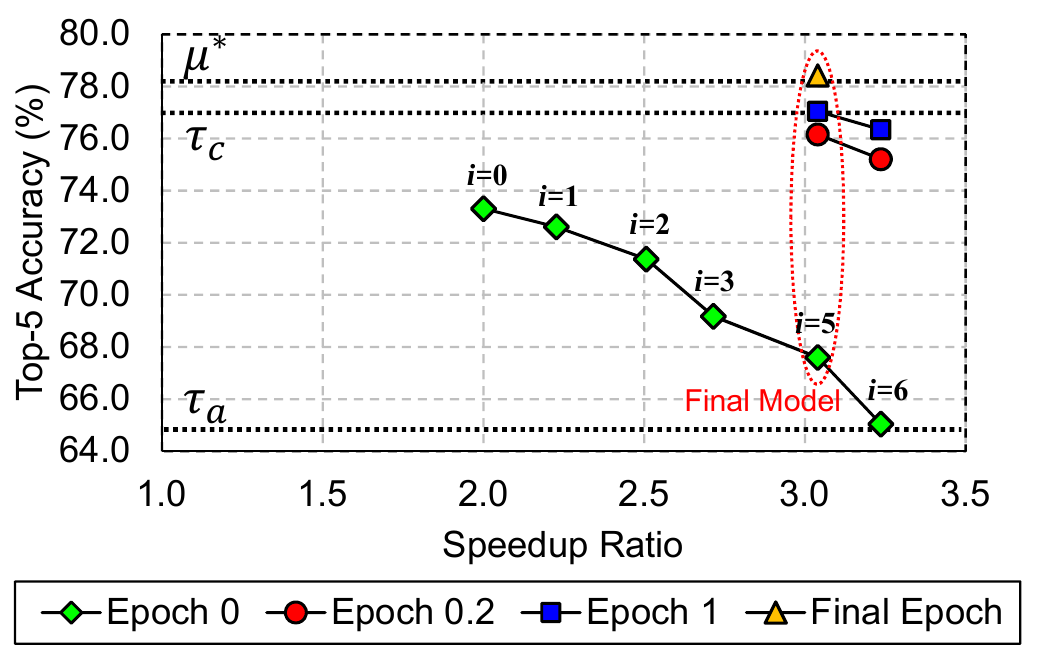}}
\end{center}
\vspace{-5.0ex}
   \caption{AlexNet results for balanced optimization (runtime and memory usage). (a) There are 8 iterations for cost reduction, where $i$ is iteration index and ${\Delta C_i}$ is the amount of cost reduction. At the gray circle, the accuracy is lower than the threshold $\tau_a$. (b) Final model satisfies all accuracy constraints }
%    ,\text{w}_3,\text{w}_4\}$ are additionally fine-tuned to compare the accuracy with $\text{w}_5$ }
% \label{fig:long}
% A rank set is the combination of elements in each vector spaces for rank
\label{fig:graph2_b}
\vspace{-3.0ex}
\end{figure}

To reduce both memory usage and runtime, we set the target cost $\epsilon_l$ as the number of operations $c_l$ and optimize the whole network including the fully-connected layers. The target accuracy$ \mu^*$ is 78.33\%\cite{kim2015compression}. 
% For each iteration, a rank set is selected from 100 candidate sets. 

% We linearly extract a group of candidate rank sets.
Two initial models with the 100\% and 50\% of total cost are fine-tuned to determine the accuracy threshold from the linear accuracy function in Fig.~\ref{fig:graph2_b}(a).  
At every iteration, the accuracies of extracted 100 candidate sets are verified without fine-tuning.
% We set the initial cost variance $\Delta C_0$=5\% and minimum cost variance $\Delta C_{min}$=1\%.
From the half-cost model, 
there are 8 iterations 
% for selecting a rank set among 100 candidate sets, and 5 models, $\{\text{w}_2, ..., \text{w}_6\}$, show the higher accuracy than threshold $\tau_a$.
for cost reduction, and 5 rank sets are updated in the list $[\textbf{R}]$ by satisfying the accuracy condition $\tau_a$. 
Since the accuracy of optimized models at $i$=(4,7,8) is lower than accuracy threshold $\tau_a$, the cost is reduced at the previous model satisfying $\tau_a$.

After cost reduction is done, the last optimized model is fine-tuned to verify the accuracy constraints at 0.2 and 1 training epoch.
In our experiment, the last model ($i$=7) does not satisfy the accuracy threshold $\tau_c$ at 1 training epoch. Therefore, the previous model ($i$=6) is fine-tuned, and it satisfies all accuracy threshold and target accuracy. 
% The accuracies of optimized models almost follow the linear accuracy function during the optimization.
% The reference model \cite{?} exploits the combination of Tucker-decomposition and VBMF based rank selection.
To recover the accuracy, we set the base learning rate as $10^{-4}$ and decrease it by a factor of 2 every 4 epochs with a batch size of 256 till 32 epochs.
% In Fig.~\ref{fig:graph2_b}, the other optimized models, $\{\text{w}_2,\text{w}_3,\text{w}_4,\text{w}_6\}$, are additionally fine-tuned to verify the linear accuracy condition. 

Compared to the combination of Tucker-decomposition \cite{tucker1966some} and VBMF based rank selection \cite{kim2015compression}, our final model shows 3.04 times speedup and 78.43\% accuracy while reducing significant memory usage as denoted in Table.~\ref{table:graph2_b}.

\subsection{Experiments with VGG-16}

\subsubsection{Runtime optimization}

\begin{table}[t]
\centering
\caption{Performance comparison. FLOPs is computed including fully-connected layers}
\vspace{-2.0ex}
\renewcommand{\tabcolsep}{0.12cm} % Default value: 1
\begin{tabular}{@{}cccccccc@{}}
\toprule
\begin{tabular}[c]{@{}c@{}}VGG-16\vspace{\tabgab}\end{tabular} & \multicolumn{1}{c}{\begin{tabular}[c]{@{}c@{}} FLOPs\end{tabular}} & \multicolumn{1}{c}{\begin{tabular}[c]{@{}c@{}} Top-5 Acc.\end{tabular}}  & \multicolumn{1}{c}{\begin{tabular}[c]{@{}c@{}} Target Acc.\end{tabular}} & \multicolumn{1}{c}{\begin{tabular}[c]{@{}c@{}} Decomp. \end{tabular}} & \multicolumn{1}{c}{\begin{tabular}[c]{@{}c@{}} Rank Sel.\end{tabular}}   \\ \midrule
% X. Zhang ($\times3$)~\cite{zhang2016accelerating}  &
\begin{tabular}[c]{@{}c@{}}{X. Zhang}\vspace{\tabgab}\\ {($\times3$)~\cite{zhang2016accelerating}}\end{tabular} &
% 4764 M ($\times$3.25)  & 
\begin{tabular}[c]{@{}c@{}}{4764 M}\vspace{\tabgab}\\ {($\times$3.26)}\end{tabular} &
89.9\% & - & 
\begin{tabular}[c]{@{}c@{}}{\textit{3D}: Asym.}\vspace{\tabgab}\\ {(3-level)}\end{tabular} &
\begin{tabular}[c]{@{}c@{}}{layer-wise}\vspace{\tabgab}\\ {search}\end{tabular} &
\vspace{\cellgabb}\\
% \textbf{{Ours} (fine-tuned)}  & 
\begin{tabular}[c]{@{}c@{}}{\textbf{Ours}}\vspace{\tabgab}\\ {(final model)}\end{tabular} &
% \textbf{3775 M ($\times$4.10)}  & 
\begin{tabular}[c]{@{}c@{}}\textbf{3837 M}\vspace{\tabgab}\\ \textbf{($\times$4.03)}\end{tabular} &
90.0\%  & 89.9\%\cite{simonyan2014very} &
\begin{tabular}[c]{@{}c@{}}{SVD}\vspace{\tabgab}\\ {(2-level)}\end{tabular} &
% SVD (2-level) &
\begin{tabular}[c]{@{}c@{}}{model-wise}\vspace{\tabgab}\\ {rank search}\end{tabular} &
\\ \bottomrule
\end{tabular}
\label{table:graph3_a}
\vspace{-3.0ex}
\end{table}
% \begin{tabular}[c]{@{}c@{}}{Model-wise}\vspace{\tabgab}\\ {Search}\end{tabular} & 

% graph3 ------------------------------
\begin{figure}[t]
\begin{center}
\subfigure[Iterative Cost Reduction]{\includegraphics[width=0.36\linewidth]{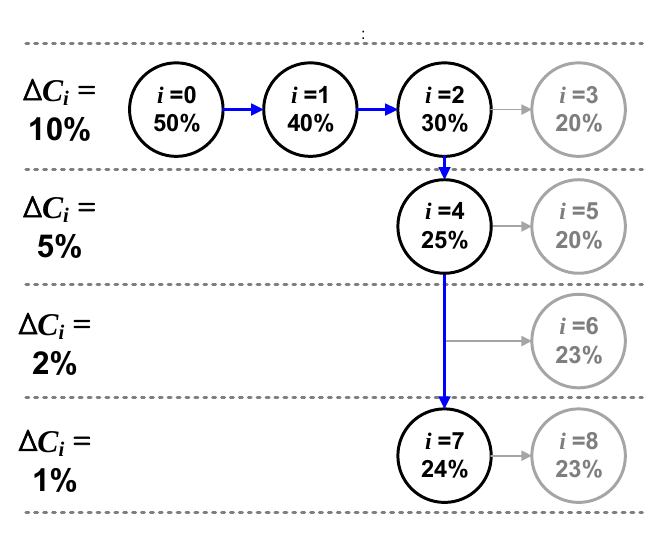}}
\subfigure[Accuracy of optimized models]{\includegraphics[width=0.5\linewidth]{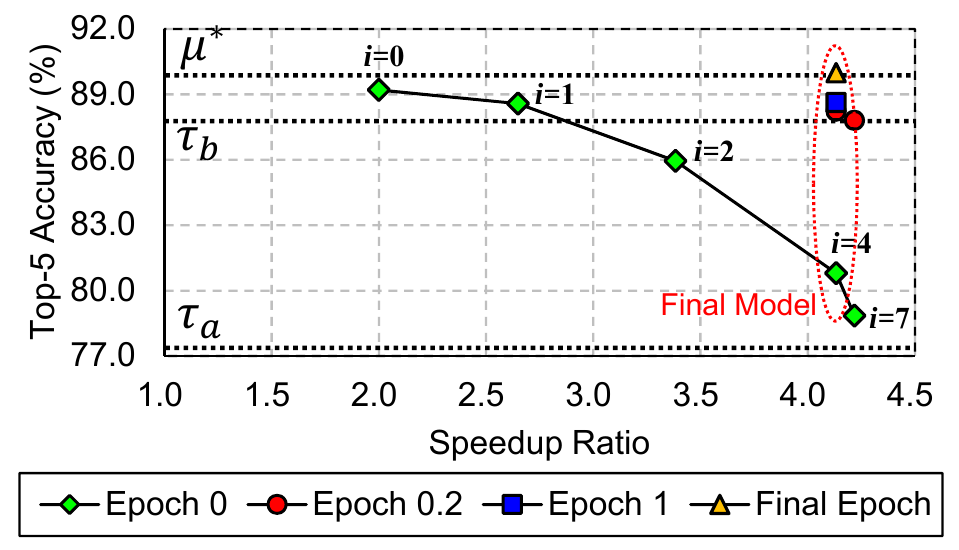}} 
\vspace{-4.0ex}
\end{center}
   \caption{VGG-16 result for runtime optimization. (a) There are 8 iterations for cost reduction, where $i$ is iteration index and ${\Delta C_i}$ is the amount of cost reduction. At the gray circle, the accuracy is lower than the threshold $\tau_a$. (b) Final model satisfies all accuracy constraints}
\label{fig:graph3}
\vspace{-1.0ex}
\end{figure}

While VGG-16 model has a high top-5 classification accuracy of 89.9\%, the weight parameters are highly redundant \cite{zhang2016accelerating}. To maintain the original accuracy, we optimize the runtime of VGG-16 and compare our optimization results with the recent works \cite{zhang2016accelerating,He_2017_ICCV}.
% which uses the layer-wise rank selection and \cite{He_2017_ICCV}.
The target accuracy is 89.9\%~\cite{simonyan2014very}.
 
First, we fine-tune two models with 50\% and 25\% of the maximum cost to determine the accuracy threshold.
From the half-cost model without fine-tuning, the rank selection from 200 candidate sets is repeated for 8 iterations, 
and 4 rank sets are updated in the list $[\textbf{R}]$ by satisfying the accuracy condition $\tau_a$.
%4 models, $\{\text{v}_2, ... ,\text{v}_5\}$, show higher accuracy than $\tau_a$.

The brief fine-tuning of last selected model ($i$=7) is terminated at 0.2 epoch, since the test accuracy is lower than the threshold $\tau_b$.
Therefore, we fine-tune the previous model ($i$=6), and check the accuracy requirement for all the thresholds. 
For fine-tuning, we set the base learning rate as $10^{-5}$ and decrease it by a factor of 10 every 4 epochs with a batch size of 16 up to 8 epochs. % due to the lack of GPU memory.  

\begin{table*}[t] \small
\centering
\caption{Performance comparison of state-of-the-art work}
\vspace{-2.0ex}
\renewcommand{\tabcolsep}{0.2cm} % Default value: 1
\begin{tabular}{@{}cccc@{}}
\toprule
\multirow{2}{*}{\begin{tabular}[c]{@{}c@{}}Models\vspace{\tabgab}\\ (VGG-16)\vspace{-1.0ex}\end{tabular}} & \multicolumn{2}{c}{Top-5 Accuracy} & 
\multirow{2}{*}{\begin{tabular}[c]{@{}c@{}} Optimization Method\vspace{\tabgab}\end{tabular}}   \\ 
\cmidrule(l){2-3} & 
\multicolumn{1}{c}{w/ FT}   & 
\multicolumn{1}{c}{w/o FT}   & 
% & Kernel Optimization                 & Rank Selection                           
\\ \midrule
% X. Zhang ($\times3$)~\cite{zhang2016accelerating}     &
% \begin{tabular}[c]{@{}c@{}}{\textit{3C}: C-P ($\times4$) ~\cite{He_2017_ICCV}}\end{tabular} &
% \begin{tabular}[c]{@{}r@{}}89.9\%\vspace{\tabgab}\end{tabular} & \begin{tabular}[c]{@{}r@{}}-\end{tabular} &
% handcraft 
% \vspace{\cellgab}\\
% Y. He~\cite{He_2017_ICCV}   & 
\begin{tabular}[c]{@{}c@{}}{ADC ($\times4$) ~\cite{ADC}}\end{tabular} &
\begin{tabular}[c]{@{}r@{}}-\end{tabular} & \begin{tabular}[c]{@{}r@{}}80.7\%\end{tabular}  &
reinforcement learning 
\vspace{\cellgab}\\
% \textbf{{Ours ($\text{v}_{4}$)}}     
\begin{tabular}[c]{@{}c@{}}{\textbf{Ours} \textbf{($\times4.03$)}}\end{tabular} &
\begin{tabular}[c]{@{}r@{}}\textbf{90.0\%}\end{tabular}  & 
\begin{tabular}[c]{@{}r@{}}\textbf{80.8\%}\end{tabular} &
rank searching algorithm
\\ \bottomrule
\end{tabular}
\label{table:graph3_b}
% \vspace{-2.0ex}
\end{table*}

% Table 1 ---------------------------- 
% \clearpage

% \usepackage{booktabs}
% \usepackage{multirow}

\begin{table*}[t] \small
\centering
\caption{Comparison of CPU and GPU processing time for forward and backward pass. Forward pass is related to inference and backward pass is related to training. The runtime is for a single view (\textit{the performance results are based on our implementation of \cite{zhang2016accelerating,He_2017_ICCV}})}
\vspace{-2.0ex}
\renewcommand{\tabcolsep}{0.2cm} % Default value: 1
\begin{tabular}{@{}cccccc@{}}
\toprule
\multirow{2}{*}{\begin{tabular}[c]{@{}c@{}}Models\vspace{\tabgab}\\ (VGG-16)\vspace{-1.0ex}\end{tabular}} & \multicolumn{2}{c}{Forward Speed} & 
\multicolumn{2}{c}{Backward Speed} & 
\multirow{2}{*}{\begin{tabular}[c]{@{}c@{}} Decomposed\vspace{\tabgab}\\ Kernel Layers\vspace{\tabgab}\end{tabular}}   \\ 
\cmidrule(l){2-5} & 
\multicolumn{1}{c}{CPU {[}ms{]}}   & 
\multicolumn{1}{c}{GPU {[}ms{]}}   & 
\multicolumn{1}{c}{CPU {[}ms{]}}   & 
\multicolumn{1}{c}{GPU {[}ms{]}}                        
% & Kernel Optimization                 & Rank Selection                           
\\ \midrule
% X. Zhang ($\times3$)~\cite{zhang2016accelerating}     &
\begin{tabular}[c]{@{}c@{}}{X. Zhang}\vspace{\tabgab}\\ {($\times3$)~\cite{zhang2016accelerating}}\end{tabular} &
\begin{tabular}[c]{@{}r@{}}1088.42\vspace{\tabgab}\\ ($\times$2.14)\end{tabular} & \begin{tabular}[c]{@{}r@{}}9.98\vspace{\tabgab}\\ ($\times$1.14)\end{tabular} & \begin{tabular}[c]{@{}r@{}}1005.15\vspace{\tabgab}\\ ($\times$1.87)\end{tabular} & \begin{tabular}[c]{@{}r@{}}14.15\vspace{\tabgab}\\ ($\times$1.28)\end{tabular} &
37 (34 Conv, 3 FC)
\vspace{\cellgab}\\
% Y. He~\cite{He_2017_ICCV}   & 
\begin{tabular}[c]{@{}c@{}}{Y. He~\cite{He_2017_ICCV}}\vspace{\tabgab}\\ {(\textit{3C}: C-P)}\end{tabular} &
\begin{tabular}[c]{@{}r@{}}968.31\vspace{\tabgab}\\ ($\times$2.40)\end{tabular} & \begin{tabular}[c]{@{}r@{}}8.62\vspace{\tabgab}\\ ($\times$1.31)\end{tabular} & \begin{tabular}[c]{@{}r@{}}892.68\vspace{\tabgab}\\ ($\times$2.49)\end{tabular} & \begin{tabular}[c]{@{}r@{}}12.94\vspace{\tabgab}\\ ($\times$1.40)\end{tabular} & 
37 (34 Conv, 3 FC)
\vspace{\cellgab}\\
% \textbf{{Ours ($\text{v}_{4}$)}}     
\begin{tabular}[c]{@{}c@{}}{\textbf{Ours}}\vspace{\tabgab}\\ {(final model)}\end{tabular}
& \begin{tabular}[c]{@{}r@{}}\textbf{824.35}\vspace{\tabgab}\\ \textbf{($\times$2.82)}\end{tabular}  & \begin{tabular}[c]{@{}r@{}}\textbf{8.39}\vspace{\tabgab}\\ \textbf{($\times$1.35)}\end{tabular} & \begin{tabular}[c]{@{}r@{}}\textbf{761.24}\vspace{\tabgab}\\ \textbf{($\times$2.92)}\end{tabular}  & \begin{tabular}[c]{@{}r@{}}\textbf{12.11}\vspace{\tabgab}\\ \textbf{($\times$1.49)}\end{tabular} & 
29 (26 Conv, 3 FC)
\\ \bottomrule
\end{tabular}
\label{table:graph3_b}
\vspace{-2.0ex}
\end{table*}

Finally, the optimized model shows 90.0\% accuracy and achieves a 4.03 times speedup as shown in Table.~\ref{table:graph3_a} and Fig.~\ref{fig:graph3}(b). 
Our method using the SVD and model-wise rank selection shows better performance than the combination of asymmetric reconstruction and layer-wise rank selection \cite{zhang2016accelerating}.
Also, the optimization performance of our algorithm is comparable to the state-of-the-art works, ADC~\cite{ADC} for spatial decomposition. 
We expect the proposed algorithm can be effectively applied to other network acceleration algorithms such as asymmetric reconstruction~\cite{zhang2016accelerating} and channel-pruning \cite{He_2017_ICCV} to achieve higher accuracy.

% In our optimization, we set the minimum cost reduction $\Delta{C}_min=5\%$ as the termination condition.
% Therefore when $\Delta{C}_min$ is set to a smaller value, 

% Training policy
% At the end of optimization, we additionally fine-tune the previous models, $\{\text{v}_2, \text{v}_3, \text{v}_4\}$, to verify the accuracy.
% The optimized model $\text{v}_4$ achieves \textbf{90.0\%} accuracy for a speedup ratio of $\times4.02$, which is 0.1\% higher accuracy than the state-of-the-art of $\times4$ acceleration model in \cite{zhang2016accelerating,He_2017_ICCV}.

\subsubsection{Processing Time}

%(Hyeji) 최적화된 network model 의 총 layer 수가 더 작음을 설명

We perform the runtime benchmark of optimized models on Nvidia GTX 1080 GPU and Intel Zeon E5-2620 CPU.
We use the standard library in \textit{Caffe} to measure the processing time of forward and backward pass.
To fairly compare the precessing time, we implement the CNN models based on the network specification of X. Zhang \textit{\textit{et al.}} ($\times3$) \cite{zhang2016accelerating} and Y. He \textit{\textit{et al.}} \cite{He_2017_ICCV}.  
Also, we compare the runtime performance at almost same accuracy, since there is a trade-off between accuracy and computation time.
Table.~\ref{table:graph3_b} shows the results of GPU and CPU processing times for a single view.
Our method uses 2-level decomposition based on truncated SVD
% channel decomposition for first convolutional layer and spatial decomposition for other layers
, whereas \cite{zhang2016accelerating,He_2017_ICCV} use 3-level spatial and channel decomposition. 
Therefore, we can implement a shorter CNN model, and it provides faster processing time than \cite{zhang2016accelerating,He_2017_ICCV} in terms of inference (i.e. forward pass) and training (i.e. forward and backward pass).

% \subsubsection{Optimization Time} 

% % We perform the runtime benchmark of optimized models on Nvidia GTX 1080 GPU and Intel Zeon E5-2620 CPU.
% % We use the standard library in \textit{Caffe} to measure the processing time of forward and backward pass.
% % To fairly compare the precessing time, we implement the CNN models based on the network specification of X. Zhang \textit{\textit{et al.}} ($\times3$) \cite{zhang2016accelerating}.
% % Also, we compare the runtime performance at almost same accuracy, since there is a trade-off between accuracy and computation time.
% % Also, we compare the runtime performance at almost same accuracy, since there is a trade-off between accuracy and computation time.

% % The proposed rank selection method is 
% Compared to the layer-wise rank selection in \cite{zhang2016accelerating}, 
% Compared to the recent works \cite{zhang2016accelerating,He_2017_ICCV} for the decomposition, our optimization method is concentrated on the rank selection. 

% ============================================================================

%%%%%%%%%%%%%%%%%%%%%%%%%%%%%%%%%%%%%%%%%%%%%%%%%%%%%%%%%%%%%%%%%%%%%%%%%%%%%%%%%%%%%%%%%%%%%

\section{Conclusion}
In this paper, we propose a model-wise rank selection algorithm that minimizes the CNN complexity while satisfying the target accuracy.
We define the rank selection problem as the combinatorial optimization, and propose the space limitation parameters to reduce the search space and obtain an optimal solution.
Also, we define a linearly-approximated accuracy function to predict the recovered accuracy in the rank selection stage. 
% We evaluate the performance with runtime and balanced optimization, which are .
From experiments on AlexNet and VGG-16, we show that the proposed optimal rank selection algorithm successfully satisfies the target accuracy and provides a faster CNN model while maintaining the same accuracy.

% \clearpage

\bibliographystyle{unsrt}
% \textbf{References}
\bibliography{egbib}

\end{document}